\documentclass[10pt,twocolumn,letterpaper]{article}
\usepackage{iccv}
\usepackage{times}
\usepackage{epsfig}
\usepackage{graphicx}
\usepackage{amsmath}
\usepackage{amssymb}
\usepackage{graphicx}
\usepackage{subcaption}
\usepackage{multirow}
\usepackage{multicol}
\usepackage{floatrow}
\usepackage{algorithm}
\usepackage{algorithmic}
\usepackage{booktabs}
\usepackage{mathrsfs}
\usepackage{colortbl}  %彩色表格需要加载的宏包
\usepackage{xcolor, eucal}
\usepackage{array}   %对表列和表格线的设置需要用到array宏包
\definecolor{gray0}{gray}{0.9}

\renewcommand{\Sigma}{\mathfrak{S}}

\def\eqref#1{equation~\ref{#1}}

\def\1{\bm{1}}

\DeclareMathAlphabet{\mathsfit}{\encodingdefault}{\sfdefault}{m}{sl}
\SetMathAlphabet{\mathsfit}{bold}{\encodingdefault}{\sfdefault}{bx}{n}

\newcommand{\Ours}{\textup{GenPromp}\xspace}

\usepackage{array}
\newcolumntype{I}{!{\vrule width 3pt}}
\newlength\savedwidth

\newlength\savewidth

\usepackage{tablefootnote}
\usepackage{tabulary,multirow,overpic,xcolor,subfloat}

\definecolor{linkcolor}{HTML}{ED1C24}

\newcommand{\app}{\raise.17ex\hbox{$\scriptstyle\sim$}}

\newcolumntype{x}[1]{>{\centering\arraybackslash}p{#1pt}}
\newcolumntype{y}[1]{>{\raggedright\arraybackslash}p{#1pt}}

\definecolor{Gray}{gray}{0.5}
\newcommand{\tablestyle}[2]{\setlength{\tabcolsep}{#1}\renewcommand{\arraystretch}{#2}\centering\footnotesize}
\makeatletter\renewcommand\paragraph{\@startsection{paragraph}{4}{\z@}
  {.5em \@plus1ex \@minus.2ex}{-.5em}{\normalfont\normalsize\bfseries}}\makeatother

\def\eg{\emph{e.g.}}

\def\ie{\emph{i.e.}}

\usepackage[pagebackref=true,breaklinks=true,letterpaper=true,colorlinks,bookmarks=false]{hyperref}

\iccvfinalcopy % *** Uncomment this line for the final submission

\def\iccvPaperID{10524} % *** Enter the ICCV Paper ID here

\ificcvfinal\fi

\begin{document}

%%%%%%%%% TITLE
\title{Generative Prompt Model for Weakly Supervised Object Localization}

\author{Yuzhong Zhao$^1$,
~~
Qixiang Ye$^1$,
~~
Weijia Wu$^2$,
~~
Chunhua Shen$^2$,
~~
Fang Wan$^1$\thanks{Corresponding author}
\\[0.205cm]
$^1$ University of Chinese Academy of Sciences~~~
$^2$ Zhejiang University\\
}

% \and

\maketitle
\ificcvfinal\thispagestyle{empty}\fi

%abstract
\begin{abstract}
Weakly supervised object localization (WSOL) remains challenging when learning object localization models from image category labels.
%问题
Conventional methods that discriminatively train activation models ignore representative yet less discriminative object parts.
In this study, we propose a generative prompt model (GenPromp), defining the first generative pipeline to localize less discriminative object parts by formulating WSOL as a conditional image denoising procedure.
During training, GenPromp converts image category labels to learnable prompt embeddings which are fed to a generative model to conditionally recover the input image with noise and learn representative embeddings.
During inference, GenPromp combines the representative embeddings with discriminative embeddings (queried from an off-the-shelf vision-language model) for both representative and discriminative capacity.
The combined embeddings are finally used to generate multi-scale high-quality attention maps, which facilitate localizing full object extent.
Experiments on CUB-200-2011 and ILSVRC show that GenPromp respectively outperforms the best discriminative models by 5.2\% and 5.6\% (Top-1 Loc), setting a solid baseline for WSOL with the generative model. Code is available at \url{https://github.com/callsys/GenPromp}.
\end{abstract}

\begin{figure}[!t]
	\includegraphics[width=0.99\linewidth]{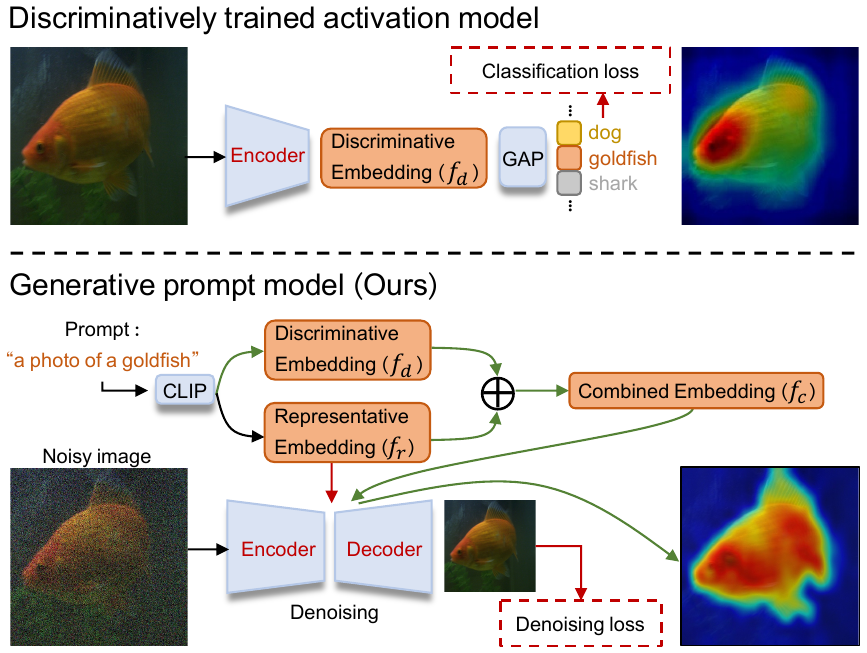}
	\caption{\textbf{Comparison of our generative prompt model (\Ours) with the discriminatively trained activation model.} \Ours aims to localize less discriminative object parts by formulating WSOL as a conditional image denoising procedure. \textcolor[rgb]{0.75,0,0}{Red}, \textcolor[rgb]{0.33,0.51,0.21}{green}, and black arrows respectively denote information propagation during training, inference, and training $\&$ inference. $f_d$, $f_r$, and $f_c$ denote the discriminative, representative, and combined embeddings.}
\label{fig:motivation}
\end{figure}

\section{Introduction}
Weakly supervised object localization (WSOL) is a challenging task when provided with image category supervision but required to learn object localization models. As a pioneered WSOL method, Class Activation Map (CAM)~\cite{DBLP:conf/cvpr/ZhouKLOT16} defines global average pooling (GAP) to generate semantic-aware localization maps based on a discriminatively trained activation model. Such a fundamental method, however, suffers from partial object activation while often missing full object extent, Fig.~\ref{fig:motivation}(upper). The nature behind the phenomenon is that discriminative models are born to pursue compact yet discriminative features while ignoring representative ones~\cite{DBLP:conf/eccv/BaeNK20,DBLP:conf/iccv/GaoWPP0HZY21}.   

Many efforts have been proposed to alleviate the partial activation issue by introducing spatial regularization terms~\cite{DBLP:conf/cvpr/KimKLKY22,DBLP:conf/eccv/LuJXSZ020, DBLP:conf/cvpr/MaiYL20,DBLP:conf/cvpr/WuZC22, DBLP:conf/iccv/XueLWJJY19, gong2022curiosity, DBLP:conf/iccv/YunHCOYC19, DBLP:conf/cvpr/ZhangWF0H18, DBLP:conf/eccv/ZhangWKYH18}, auxiliary localization modules~\cite{DBLP:conf/eccv/LuJXSZ020,DBLP:conf/iccv/MengZ00021,DBLP:conf/iccv/XieLZJLS21,DBLP:conf/cvpr/XieXCHZS22}, or adversarial erasing strategies~\cite{DBLP:journals/pami/ChoeLS21,DBLP:conf/cvpr/ChoeS19,DBLP:conf/cvpr/MaiYL20,DBLP:conf/cvpr/ZhangWF0H18}. Nevertheless, the fundamental challenge about how to use a discriminatively trained classification model to generate precise object locations remains.

\begin{figure*}[t]
	\includegraphics[width=1\linewidth]{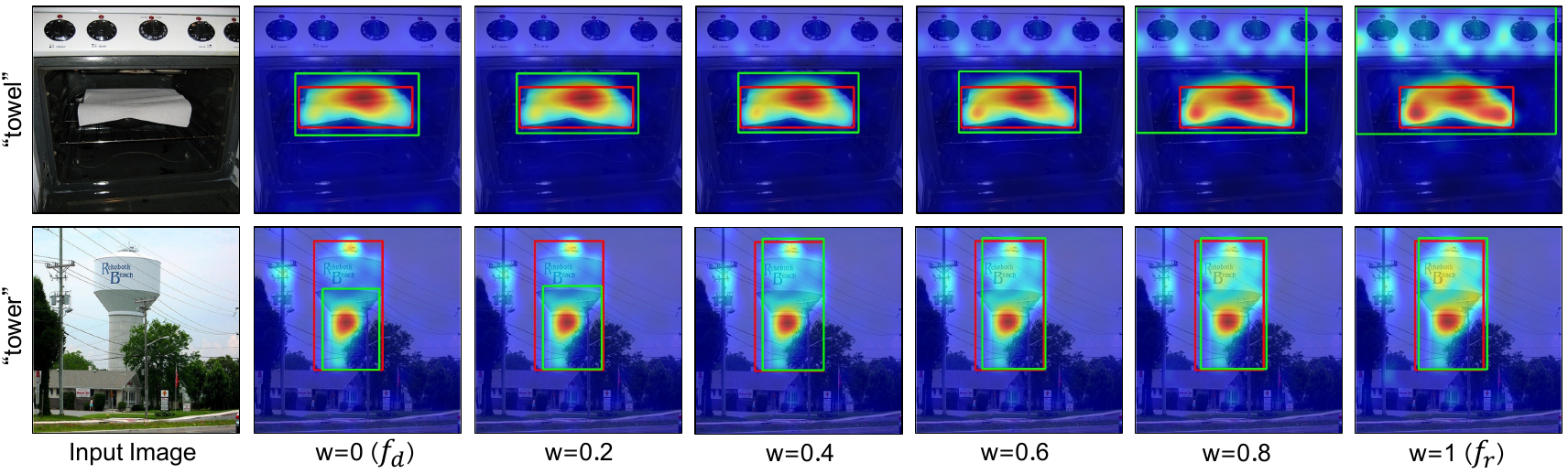}
	\caption{\textbf{Activation maps and localization results using discriminative and representative embeddings.} A proper combination ($w$=0.6) of discriminative embeddings $f_d$ with representative embedding $f_r$ as the prompt produces precise activation maps and good WSOL results (green boxes). }
\label{fig:viz_token}
\end{figure*}

In this study, we propose a generative prompt model (\Ours), Fig.~\ref{fig:motivation}(lower), which formulates WSOL as a conditional image denoising procedure, solving the fundamental partial object activation problem in a new and systematic way. 
During training, for each category (\eg{\texttt{goldfish} in Fig.~\ref{fig:motivation}}) in the predetermined category set, GenPromp converts each category label to a learnable prompt embedding ($f_r$) through a pre-trained vision-language model (CLIP)~\cite{DBLP:conf/icml/RadfordKHRGASAM21}.
The learnable prompt embedding is then fed to a transformer encoder-decoder to conditionally recover the noisy input image.
% %
Through multi-level denoising, the representative features of input images are back-propagated from the transformer decoder to the learnable prompt embedding, which is updated to the representative embedding $f_r$.

During inference, GenPromp linearly combines learned representative embeddings ($f_r$) with discriminative embeddings ($f_d$) to obtain both object generative and discrimination capability, Fig.~\ref{fig:viz_token}. 
$f_d$ is queried from a pre-trained vision-language model (CLIP), which incorporates the correspondence between text (\eg{category labels}) with vision feature embeddings. The combined embedding ($f_c$) is used to generate attention maps at multiple levels and timestamps, which are aggregated to object activation maps through a post-processing strategy. 
On CUB-200-2011~\cite{WahCUB_200_2011} and ImageNet-1K~\cite{DBLP:journals/ijcv/RussakovskyDSKS15}, \Ours respectively outperforms the best discriminative models by 5.2\% (87.0\% $vs.$ 81.8\%~\cite{DBLP:conf/cvpr/XieXCHZS22}) and 5.6\% (65.2\% $vs.$ 59.6\%~\cite{DBLP:conf/cvpr/XieXCHZS22}) in Top-1 Loc. 

The contributions of this study include:
\begin{itemize}
    \item We propose a generative prompt model (\Ours), providing a systematic way to solve the inconsistency between the discriminative models with the generative localization targets by formulating a conditional image denoising procedure.
    
    \item We propose to query discriminative embeddings from an off-the-shelf vision-language model using image labels as input. Combining the discriminative embeddings with the generative prompt model facilitates localizing objects while depressing backgrounds. 
    
    \item \Ours significantly outperforms its discriminative counterparts on commonly used benchmarks, setting a solid baseline for WSOL with generative models.
    
\end{itemize}

\section{Related Work}

\textbf{Weakly Supervised Object Localization.} 
As a simple-yet-effective method, CAM~\cite{DBLP:conf/cvpr/ZhouKLOT16} localizes objects by introducing global average pooling (GAP) to an image classification network. CAM is also extended from WSOL to weakly supervised  detection~\cite{DBLP:journals/tip/ChengYGGH20,DBLP:journals/pami/WanWHJY19,DBLP:journals/pami/ZhangZYH22} and segmentation ~\cite{DBLP:conf/cvpr/ChenTZ0D22,DBLP:conf/cvpr/WangZKSC20,DBLP:conf/cvpr/XieXCHZS22,DBLP:conf/iccv/ZhangGZD21}. However, CAM suffers from partial activation, $i.e.$, activating the most discriminative parts instead of full object extent. The reason lies in the inconsistency between the discriminative models ($i.e.$, classification model) with the generative target (object localization).

To solve the partial activation problem, adversarial erasing, discrepancy learning, online localization refinement, classifier-localizer decoupling, and attention regularization methods are proposed. 
As a spatial regularization method, adversarial erasing~\cite{DBLP:journals/pami/ChoeLS21,DBLP:conf/cvpr/ChoeS19,DBLP:conf/cvpr/MaiYL20,DBLP:conf/iccv/SinghL17,DBLP:conf/iccv/YunHCOYC19,DBLP:conf/cvpr/ZhangWF0H18} online removes significantly activated regions within feature maps to drive learning the missed object parts. With a similar idea, spatial discrepancy learning~\cite{DBLP:journals/pr/GaoWYXY22,DBLP:conf/iccv/XueLWJJY19} leverages adversarial classifiers to enlarge object areas. Through classifier-localizer decoupling, PSOL~\cite{DBLP:conf/cvpr/ZhangCW20} partitions the WSOL pipeline into two parts: class-agnostic object localization and object classification. For class-agnostic localization, it uses class-agnostic methods to generate noisy pseudo annotations and then perform bounding box regression on them without class labels. While online refinement of low-level features improves activation maps for WSOL~\cite{DBLP:conf/iccv/XieLZJLS21}, BAS~\cite{DBLP:conf/cvpr/WuZC22} specifies a background activation suppression strategy to assist the learning of WSOL models. C$^2$AM~\cite{DBLP:conf/cvpr/XieXCHZS22} generates class-agnostic activation maps using contrastive learning without category label supervision. FAM~\cite{DBLP:conf/iccv/MengZ00021} optimizes the object localizer and classifiers jointly via object-aware and part-aware attention modules. TS-CAM~\cite{DBLP:conf/iccv/GaoWPP0HZY21} and LCTR~\cite{DBLP:conf/aaai/ChenWWJSTWZC22} utilize the attention maps defined on long-range feature dependency of transformers to localize objects.

Despite the progress, existing methods typically ignore the fundamental challenge of WSOL, $i.e.$  discriminative models are required to perform representative (localization) tasks. 
% The inconsistency between the trained models with the prediction targets remains.
%

\textbf{Vision-Language Models.} 
Vision-language models have demonstrated increasing importance for vision tasks. In the early years, great efforts are paid to label image-text pairs which are important for vision-language model training~\cite{DBLP:conf/acl/ChenD11,DBLP:conf/cvpr/DasXDC13,DBLP:journals/tacl/RegneriRWTSP13,DBLP:conf/cvpr/XuMYR16}.
In recent years, the born-ed association relations of image and text on the Web facilitate collecting a massive quantity of image-text pairs~\cite{DBLP:conf/cvpr/ChangpinyoSDS21,DBLP:conf/iccv/MiechZATLS19,DBLP:conf/icml/RadfordKHRGASAM21,DBLP:journals/corr/abs-2210-08402,DBLP:journals/corr/abs-2111-02114}, which requires much lower annotation cost compared to those manually annotated datasets (See supplementary for details).
Such image-text pairs enable building the association between image category labels and visual feature embedding, which is the foundation of this study.
Based on the massive quantity of image-text pairs, we pre-train two components of \Ours (\ie{Stable diffusion~\cite{DBLP:conf/cvpr/RombachBLEO22}, CLIP~\cite{DBLP:conf/icml/RadfordKHRGASAM21}}) with two large image-text pair datasets. In specific, we pre-train the image denoising model using LAION-5B~\cite{DBLP:journals/corr/abs-2210-08402} and the CLIP model on WIT ~\cite{DBLP:conf/icml/RadfordKHRGASAM21}.

\textbf{Generative Vision Models.}
As the foundation of most generative vision models, GAN~\cite{DBLP:journals/cacm/GoodfellowPMXWO20} defines an adversarial training process, where it simultaneously trains a generative model and a discriminative model.
Many efforts have been made to improve GAN such as better optimization strategies~\cite{DBLP:journals/corr/ArjovskyCB17,DBLP:conf/nips/GulrajaniAADC17,DBLP:conf/iccv/MaoLXLWS17,DBLP:conf/iclr/MiyatoKKY18,xing2022unsupervised}, conditional image generation~\cite{DBLP:conf/cvpr/IsolaZZE17,DBLP:conf/cvpr/KarrasLA19,DBLP:journals/pami/KarrasLA21} and improved architecture~\cite{DBLP:conf/iclr/BrockDS19,DBLP:conf/iclr/KarrasALL18,DBLP:journals/corr/RadfordMC15,DBLP:conf/icml/ZhangGMO19}.
Recently, a generative vision paradigm, $i.e.$, denoising diffusion probabilistic models (DDPM)~\cite{DBLP:conf/nips/HoJA20} become popular and have the potential to surpass GAN in several vision generation tasks such as image generation~\cite{DBLP:conf/nips/DhariwalN21,DBLP:journals/corr/abs-2208-01618,DBLP:conf/icml/NicholD21,DBLP:conf/cvpr/RombachBLEO22,DBLP:journals/corr/abs-2208-12242} and image editing~\cite{DBLP:journals/corr/abs-2208-01626,DBLP:journals/corr/abs-2210-09276}.
DDPM have also been adapted to some perception tasks such as object detection~\cite{DBLP:journals/corr/abs-2211-09788} and image segmentation~\cite{DBLP:journals/corr/abs-2112-00390, DBLP:conf/iclr/BaranchukVRKB22, wu2023diffumask}, which inspires this study for WSOL.

Viewing categories labels as prompt embeddings is an important feature of \Ours, which originates from conditional image generation tasks. This follows DDPM which feeds a language description encoded by the language model (\eg{BERT~\cite{DBLP:conf/naacl/DevlinCLT19}}) to the generative model for sematic-aware image generation. 
Existing works~\cite{DBLP:journals/corr/abs-2208-01618,DBLP:journals/corr/abs-2208-01626,DBLP:journals/corr/abs-2210-09276,DBLP:journals/corr/abs-2208-12242} have proposed to learn the representative embeddings from user-provided images that contains a new object or a new style, which drive DDPM to generate synthesis images that contains that object or style.
Inspired by this learning paradigm, \Ours is proposed to learn the representative embeddings of each category, which are crucial for object localization.

\section{Preliminaries}
\label{sec:pre}
Diffusion models learn a data distribution $p(x)$ by gradually denoising a normally distributed variable, which corresponds to learning a reverse process of a Markov Chain~\cite{DBLP:conf/nips/HoJA20}. Stable diffusion~\cite{DBLP:conf/cvpr/RombachBLEO22} leverages well-trained perceptual compression model $\mathcal{E}$ to transfer the denoising diffusion process from the high-dimensional pixel space to a low-dimensional latent space, which reduces the computational burden and increases efficiency. The objective function of Stable diffusion is defined as

\begin{align}
    \mathcal{L}(\theta, f) & = \mathbb{E}_{x, \epsilon, t}\Big[||\epsilon-\epsilon_{\theta}(z_t,t,f)||^2_2\Big],\\
    z_t & = \sqrt{\overline{\alpha}_t}\mathcal{E}(x) + \sqrt{1-\overline{\alpha}_t}\epsilon, t\in T, \label{eq:zt}
\end{align}
where $\epsilon\sim N(0,1)$ is sampled from the normal distribution and $\epsilon_{\theta}(\circ,t,f)$ is the neural backbone, which is implemented as an attention-unet conditioned on time $t$ and the prompt embedding $f$. %
$\mathcal{E}$ is implemented as a VQGAN~\cite{DBLP:conf/cvpr/EsserRO21} encoder. $z_t$ is the noisy latent of the input image $x$ at time $t,t\in T = \{1,2,\cdots, 1000\}$. $\{\overline{\alpha}_t\}_{t\in T}$ denotes a set of hyperparameters that steer the levels of noise added.

\section{Generative Prompt Model}
\label{sec:gpm}
\begin{figure}[t]
	\includegraphics[width=1\linewidth]{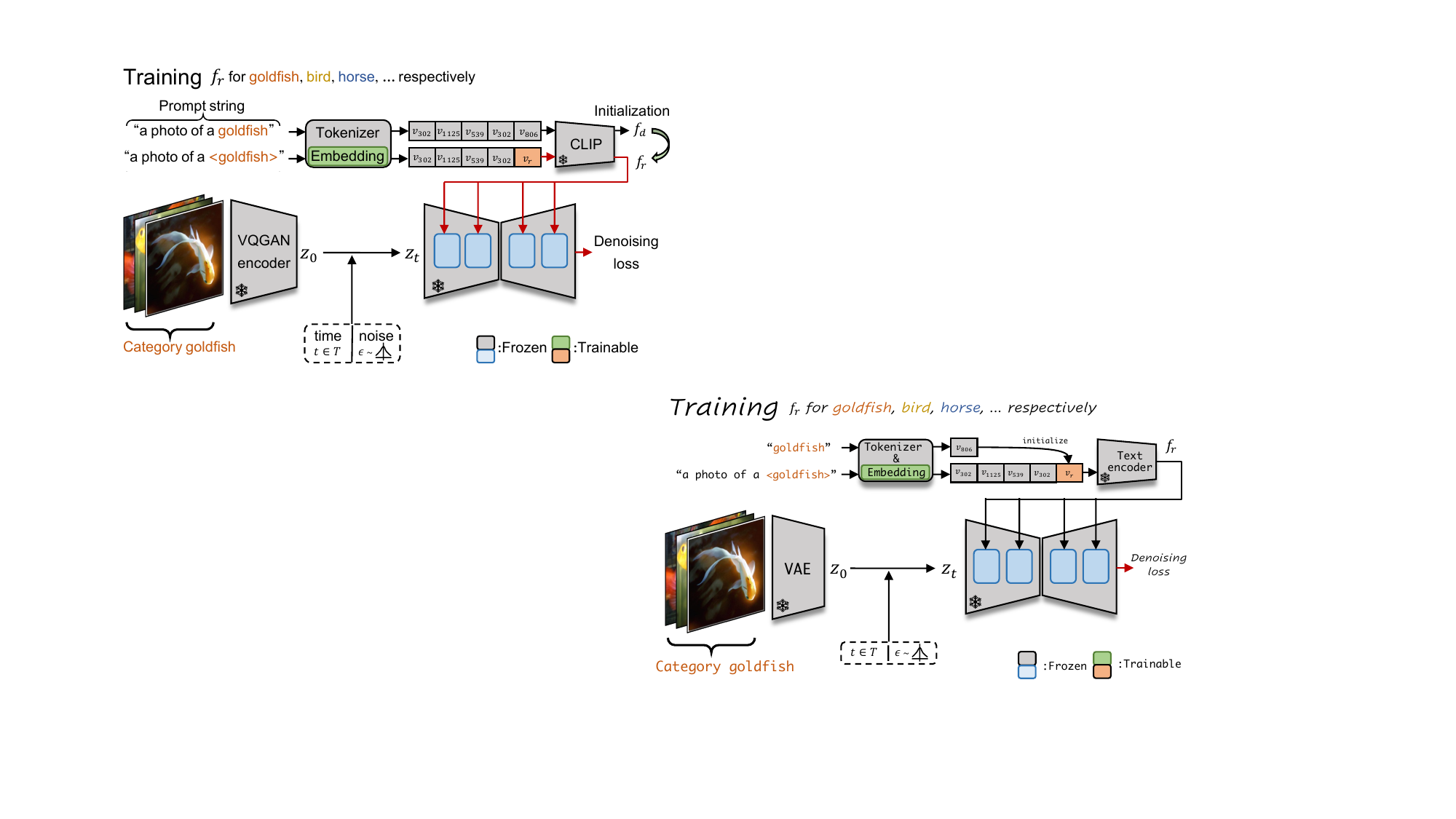}
	\caption{\textbf{Training pipeline of \Ours}. The discriminative embedding $f_d$ is queried from a CLIP model while the representative embedding $f_c$ is learned through optimizing an image denoising model.}
\label{fig:training_stage1}
\end{figure}

\begin{figure*}[t]
	\includegraphics[width=1\linewidth]{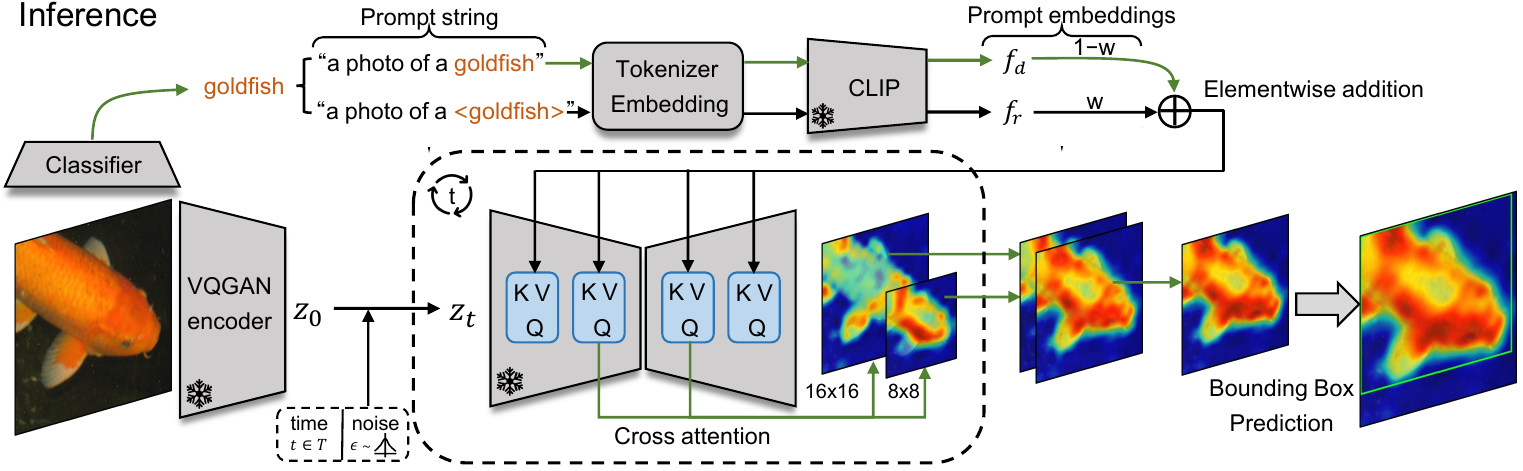}
	\caption{\textbf{Inference pipeline of \Ours.} By performing a conditional denoising procedure, \Ours produces object activation maps at multiple levels and timesteps. Two prompts (``a photo of a goldfish" and ``a photo of a $\langle$goldfish$\rangle$") are encoded to discriminative embedding $f_d$ and representative embedding $f_r$, which are combined to $f_c$. $f_c$ performs cross-attention with the features of the encoder and decoder in the image denoising model to obtain attention maps.}
\label{fig:inference}
\end{figure*}

\Ours consists of a training stage (Fig.~\ref{fig:training_stage1}) and a finetuning stage\footnote{The architecture is similar as Fig.~\ref{fig:training_stage1} but using a different training set. Please refer to the supplementary material for details.}. In the training stage, the discriminative embedding $f_d$ is queried from a pre-trained CLIP model (\textbf{Querying Discriminative Embeddings}), meanwhile a representative embedding $f_r$ is specified for each category (\textbf{Learning Representative Embedding.}). In the finetuning stage, $f_d$ and $f_r$ are used to prompt the backbone network finetuning (\textbf{Model Finetuning}). 
% The trained models and prompt embeddings are used to predict attention maps for WSOL (Fig.~\ref{fig:inference}).
The trained models and combined prompt embeddings (\textbf{Combining Embeddings}) are used to predict attention maps for WSOL (Fig.~\ref{fig:inference}).

\textbf{Querying Discriminative Embeddings.}
The embedding $f_d$ is queried from a pre-trained CLIP model~\cite{DBLP:conf/icml/RadfordKHRGASAM21}, which uses a prompt string ($p$) as input and outputs a language feature vector. 
As CLIP is pre-trained with a discriminative loss ($i.e.$, contrastive loss), $f_d$ is therefore discriminative.

During training, for a given category (\eg{\texttt{goldfish} in Fig.~\ref{fig:training_stage1}}), the prompt is obtained by filling the category label string to a template (\eg{``\texttt{a photo of a goldfish}”}). During inference, however, the image category labels are unavailable. An intuitive way is using a pre-trained classifier to predict the category of the input image. 
Unfortunately, some category labels correspond to strings of multiple words, which imply multiple tokens and multiple attention maps. Such multiple attention maps with noise would deteriorate the localization performance. To address this issue, we heuristically select the last token of the first string to initialize the prompt, which is referred to as the \textit{meta} token. For example, we select \texttt{goldfish} for category string ``\texttt{goldfish, Carassius auratus}” and select \texttt{ray} for category string ``\texttt{electric ray, crampfish, numbfish, torpedo}”. The \textit{meta} token of a category is typically the name of its superclass.

To query discriminative embedding $f_d$ from the CLIP model, the prompt $p$ is converted to an ordered list of numbers (\eg{$[$\texttt{a}$\rightarrow$302, \texttt{photo}$\rightarrow$1125, \texttt{of}$\rightarrow$539, \texttt{a}$\rightarrow$302, \texttt{goldfish}$\rightarrow$806$])$}\footnote{$\langle \texttt{BOS}\rangle$,$\langle \texttt{EOS}\rangle$ and padding tokens are omitted for clarity.}. This procedure is performed by looking up the dictionary of the $\text{Tokenizer}$. The list of numbers is used to index the embedding vectors $v(p)=[v_{302},v_{1125},v_{539},v_{302},v_{806}]$ through an index-based Embedding layer. Such embedding vectors (language vectors) are input to a pre-trained vision-language model (CLIP)~\cite{DBLP:conf/icml/RadfordKHRGASAM21} to generate the discriminative language embedding $f_d$, as
\begin{equation}
    f_d = \text{CLIP}(v(p)).\label{eq:clip}
\end{equation}

\textbf{Learning Representative Embedding.}
\label{sec:cc_token}
Using solely the discriminative embedding $f_d$, \Ours could miss representative yet less discriminative features. As an example in the second column and second row of Fig.~\ref{fig:viz_token}, the activation map of the \texttt{tower} obtained by prompting the network with $f_d$ suffers from partial object activation. The representative embedding $f_r$ is thereby introduced as a prompt learning procedure, as shown in Fig.~\ref{fig:training_stage1}. In specific, the dictionary of the \text{Tokenizer} is extended by involving a new token ($i.e.$, $\langle {\rm \texttt{goldfish}}\rangle$), which is referred to as the \textit{concept} token. The prompt with the \textit{concept} token is encoded to the representative embedding $f_r$ through the pre-trained CLIP model.

Before learning, the representative embedding $f_r$ is initialized as the discriminative embedding $f_d$. When learning representative embedding, images belonging to a same category are collected to form the training set. Each input image is encoded to the latent variable $z_0$ by the VQGAN encoder, Fig.~\ref{fig:training_stage1}. Different levels of noises are added to $z_0$ to have a noisy latent $z_t$ through Eq.~\ref{eq:zt}. The noisy latent $z_t$ is fed to the attention-unet $\epsilon_{\theta}$ to produce multi-layer feature maps $F_{l,t}$, where $l\in L$ index the encoded features of different layers. 
% Meanwhile, the image category label is used to initialize two prompts ($e.g.$, \texttt{goldfish} and $\langle\texttt{goldfish}\rangle$). Through Eq.~\ref{eq:clip}, the two prompts are respectively encoded to discriminative embedding $f_d$ and representative embedding $f_r$, which are combined to $f_c$ through Eq.~\ref{eq:couple}. 
% All the network parameters except for the embedding vector of the \textit{concept} token ($v_r$ in Fig.~\ref{fig:training_stage1}) are frozen. 
The Tokenizer Embedding layer in Fig.~\ref{fig:training_stage1} incorporate embeddings for each word/token, where the embeddings ($i.e.$, $v_r$) of the corresponding \textit{concept} tokens ($e.g.$, $\langle {\rm \texttt{goldfish}}\rangle$) are trainable.
By optimizing a denoising procedure defined by stable diffusion, the representative embedding $f_r$ is learned, as 
\begin{align}
    f_r^{*} & =\mathop{\text{argmin}}\limits_{f_r}\mathcal{L}(\theta,f_r).
\end{align}

Benefiting from the property of the generative denoising model, $f_r^{*}$ can identify the common features among objects in the training set, learning representative features that define each category. As shown in the second column of Fig.~\ref{fig:viz_token}, the activation map of \texttt{tower} obtained by prompting the network with $f_r$ activates the full object area.

\textbf{Model Finetuning.}
After obtaining the representative embeddings $f_r^{*}$ for all image categories, the backbone network (attention-unet parameterized by $\epsilon_{\theta}$) is finetuned on the WSOL dataset, as
\begin{align}
    \theta^{*} & =\mathop{\text{argmin}}\limits_{\theta}\mathcal{L}(\theta,f_d)+\mathcal{L}(\theta,f_r^{*}),
\end{align}
which further optimizes the diffusion model using both $f_d$ and $f_r^{*}$ as prompts to reduce the domain gap between the model and the target dataset. 

\textbf{Combining Embeddings.}
\label{sec:cp_token}
After model finetuning, the discriminative and representative embeddings ($f_d$ and $f_r$) are linearly combined, as
\begin{align}
    f_c & =w \cdot  f_r + (1-w) \cdot f_d,\label{eq:couple}
\end{align}
where $w\in [0,1]$ is an experimentally determined weighted factor. As shown in Fig~\ref{fig:viz_token}, for the category \texttt{towel}, a large $w$ activates background noise. Meanwhile, a small $w$ could cause partial object activation. An appropriate $w$ can balance the discriminative features and representative features of the categories, achieving the best localization performance.

\section{Weakly Supervised Object Localization}

As shown in Fig.~\ref{fig:inference}, WSOL is defined as a conditional image denoising procedure. Similar to the training procedure, each input image is encoded to the latent variable $z_0$ by the VQGAN encoder. 
%Different levels of noises are added to $z_0$ to have a noisy latent $z_t$ through Eq.~\ref{eq:zt}. 
Different levels of noise are added to $z_0$ to generate noisy latent $z_t$ through Eq.~\ref{eq:zt}.
The noisy latent $z_t$ is fed to the finetuned attention-unet $\epsilon_{\theta^{*}}$ to produce multi-layer feature maps $F_{l,t}$, where $l\in L$ index the encoded features of different layers.
Meanwhile, the input image is classified with a classifier pre-trained on the target dataset to obtain the category label, which is used to initialize two prompts (\eg{\texttt{goldfish} and $\langle\texttt{goldfish}\rangle$}).
According to Eq.~\ref{eq:clip}, the two prompts are encoded to discriminative embedding $f_d$ and representative embedding $f_r$.
The two embeddings are then combined to $f_c$, which is the condition for the image denoising model.
Through performing denoising, cross attention maps are generated by using $F_{l,t}$ as the Query vector and $f_c$ as the Key vector, as
\begin{align}
    m_{l,t} & = \text{CrossAttn}(F_{l,t},f_c),l\in L, t\in T.
\end{align}

\begin{figure*}[t]
	\includegraphics[width=1\linewidth]{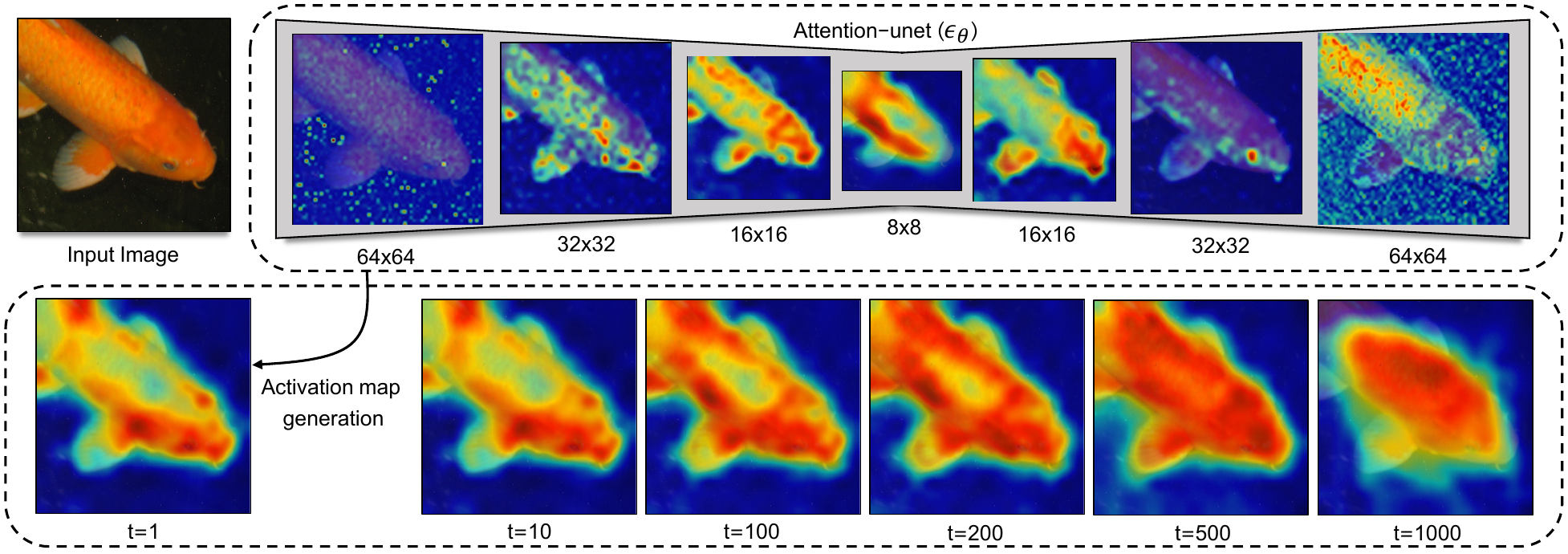}
	\caption{\textbf{Aggregation of cross attention maps.} Attention maps with respect to multiple resolutions and multiple noise levels (timesteps $t$) are aggregated to obtain the final localization map.
 % with as much as foreground while as less as background noise.
 }
\label{fig:viz_attn}
\end{figure*}

One notable capability of \Ours is to generate attention maps $\{m_{l,t}\}_{l\in L,t\in T}$ which exhibit distinct characteristics based on the specific layer $l$ at time $t$, as shown in Fig.~\ref{fig:viz_attn}. The characteristics of these attention maps can be concluded as follows: (1) Attention maps with higher resolution can provide more detailed localization clues but introduce more noise. (2) Attention maps of different layers can focus on different parts of the target object. (3) Smaller $t$ provides a less noisy background but tends to partial object activation. (4) Larger $t$ activates the target object more completely but introduces more background noise. Based on these observations, we propose to aggregate attention maps at multiple layers and timesteps to obtain a unified activation map, as
\begin{align}
    M & = \frac{1}{|L|\cdot |T|}\sum_{l\in L,t\in T}\frac{m_{l,t}}{\text{max}(m_{l,t})}\label{eq:agg}.
\end{align}
Experimentally, we find that aggregating the attention maps of spatial resolutions $8\times 8$ and $16\times 16$ at time steps $1$ and $100$ produces the best localization performance. 
A thresholding approach~\cite{DBLP:conf/cvpr/ZhangWF0H18} is then applied to predict the object locations based on the unified activation map.

\begin{table*}[t]
    \centering
    \small 
    \setlength{\tabcolsep}{1mm}
    \begin{tabular}{ r cccccccc}
\toprule
\multirow{2}{*}{Method} & \multirow{2}{*}{Loc Back.} & \multirow{2}{*}{Cls Back.} & \multicolumn{3}{c}{CUB-200-2011} & \multicolumn{3}{c}{ImageNet-1K}\tabularnewline
\cline{4-9}
 &  &  & Top-1 Loc & Top-5 Loc & GT-known Loc & Top-1 Loc & Top-5 Loc & GT-known Loc\tabularnewline
\midrule
CAM$_{\text{CVPR'16}}$~\cite{DBLP:conf/cvpr/ZhouKLOT16} & \multicolumn{2}{c}{VGG16} & 41.1 & 50.7 & 55.1 & 42.8 & 54.9 & 59.0\tabularnewline
TS-CAM$_{\text{ICCV'21}}$~\cite{DBLP:conf/iccv/GaoWPP0HZY21} & \multicolumn{2}{c}{Deit-S} & 71.3 & 83.8 & 87.7 & 53.4 & 64.3 & 67.6\tabularnewline
LCTR$_{\text{AAAI'22}}$~\cite{DBLP:conf/aaai/ChenWWJSTWZC22} & \multicolumn{2}{c}{Deit-S} & 79.2 & 89.9 & 92.4 & 56.1 & 65.8 & 68.7\tabularnewline
SCM$_{\text{ECCV'22}}$~\cite{bai2022weakly} & \multicolumn{2}{c}{Deit-S} & 76.4 & 91.6 & 96.6 & 56.1 & 66.4 & 68.8\tabularnewline
CREAM$_{\text{CVPR'22}}$~\cite{DBLP:conf/cvpr/XuHZFZZLG22} & \multicolumn{2}{c}{InceptionV3} & 71.8 & 86.4 & 90.4 & 56.1 & 66.2 & 69.0\tabularnewline
BAS$_{\text{CVPR'22}}$~\cite{DBLP:conf/cvpr/WuZC22} & \multicolumn{2}{c}{ResNet50} & 77.3 & 90.1 & 95.1 & 57.2 & 67.4 & 71.8\tabularnewline
\hline
 % &  &  &  &  &  &  &  & \tabularnewline
%
PSOL$_{\text{CVPR'20}}$~\cite{DBLP:conf/cvpr/ZhangCW20} & DenseNet161 & EfficientNet-B7 & 80.9 & 90.0 & 91.8 & 58.0 & 65.0 & 66.3\tabularnewline
C$^2$AM$_{\text{CVPR'22}}$~\cite{DBLP:conf/cvpr/XieXCHZS22} & DenseNet161 & EfficientNet-B7 & 81.8 & 91.1 & 92.9 & 59.6 & 67.1 & 68.5 \tabularnewline
\hline \rowcolor{gray0}
% GPM (Ours) & Stable Diffusion & Conformer-base-p16 &  &  &  &  &  & 74.9\tabularnewline
%
\rowcolor{gray0} GenPromp (Ours) & Stable Diffusion & EfficientNet-B7 & \textbf{87.0} & \textbf{96.1} & \textbf{98.0} & 65.1 & 73.3 & 74.9 \tabularnewline
\rowcolor{gray0} GenPromp$\dagger$ (Ours) & Stable Diffusion & EfficientNet-B7 & \textbf{87.0} & \textbf{96.1} & \textbf{98.0} & \textbf{65.2} & \textbf{73.4} & \textbf{75.0} \tabularnewline
\bottomrule
\end{tabular}
    \caption{\textbf{Performance comparison} of the proposed \Ours approach with the state-of-the-art methods on the CUB-200-2011 test set and ImageNet-1K validation set. \textit{Loc Back}. denotes the localization backbone, \textit{Cls Back}. the backbone for classification, and $\dagger$ the prompt ensemble strategy, which ensembles the localization results from multiple prompts. Please refer to the supplementary for more comparison details.}
    \label{tab:sota}
\end{table*}

\section{Experiments}
\subsection{Experimental Settings}
\textbf{Datasets.} We evaluate \Ours on two commonly used benchmarks, $i.e.$, CUB-200-2011 and ImageNet-1K. CUB-200-2011 is a fine-grained bird dataset that contains 200 categories of birds with 5994 training images and 5794 test images. ImageNet-1K is a large-scale visual recognition dataset containing 1,000 categories with 1.2 million training images and 50,000 validation images.

\textbf{Evaluation Metrics.} We follow the previous methods and use Top-1 localization accuracy (Top-1 Loc), Top-5 localization accuracy (Top-5 Loc), and GT-known localization accuracy (GT-known Loc) as the metrics. For localization, a bounding box prediction is positive when it satisfies: (1) the predicted category label is correct, and (2) the IoU between the bounding box prediction and one of the ground-truth boxes is greater than 50$\%$. GT-known indicates that it considers only the IoU constraint regardless of the classification result.

\textbf{Implementation Details.}
\Ours is implemented based the Stable Diffusion model~\cite{DBLP:conf/cvpr/RombachBLEO22}, which is pre-trained on LAION-5B~\cite{DBLP:journals/corr/abs-2210-08402}. The text encoder, $i.e.$, CLIP, is pre-trained on WIT~\cite{DBLP:conf/icml/RadfordKHRGASAM21}. 
During training, we resize the input image to 512$\times$512 and augment the training data with \text{RandomHorizontalFlip} and \text{ColorJitter}. We then optimize the network using AdamW with $\epsilon$=1$e$$-$8, $\beta_1$=0.9, $\beta_2$=0.999 and weight decay of 1$e$$-$2 on 8 RTX3090. In the training stage, we optimize \Ours for 2 epochs with learning rate 5$e$$-$5 and batch size 8 for each category in CUB-200-2011 and ImageNet-1K. In the finetuning stage, we train \Ours for 100,000 iterations with learning rate 5$e$$-$8 and batch size 128.

\begin{figure}[t]
	\includegraphics[width=1\linewidth]{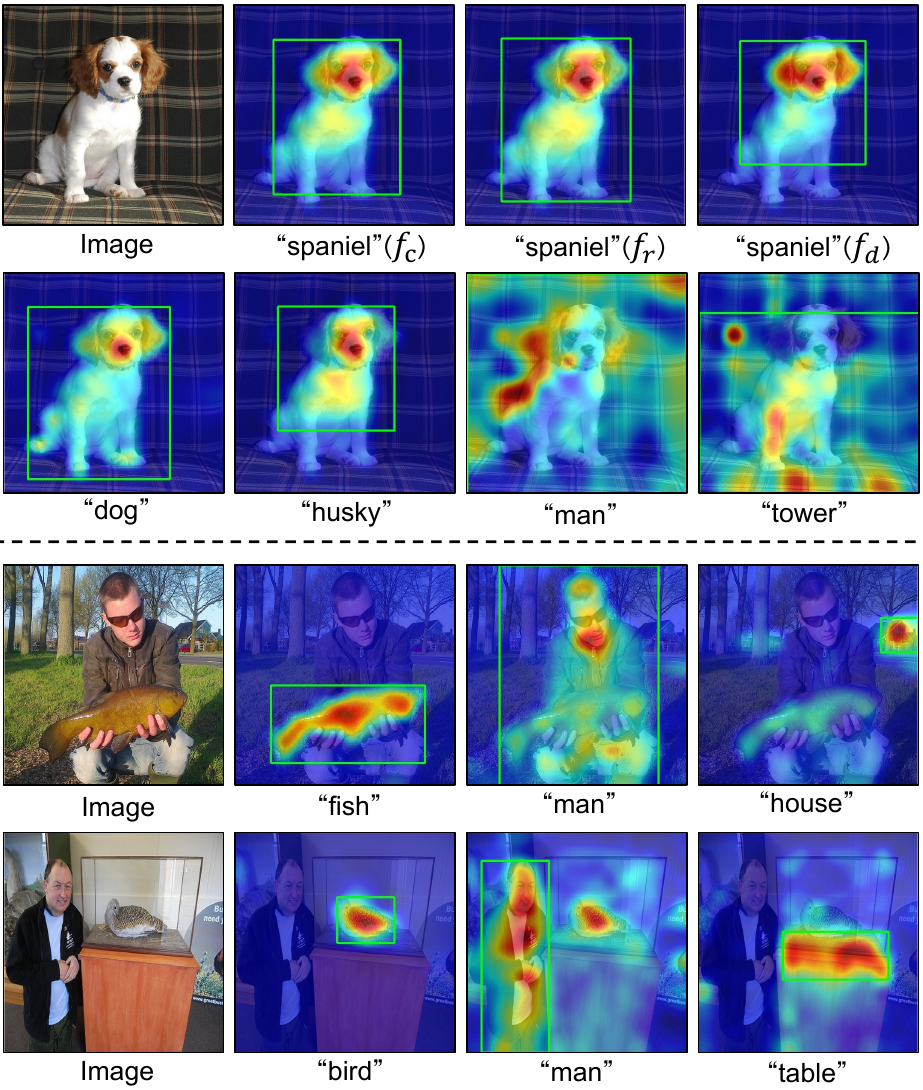}
	\caption{{Object localization results of \Ours using different prompt words.} Words belonging to the same superclass activate the same object (upper), and words from different superclasses tend to activate different regions (upper and lower).}
\label{fig:viz_query}
\end{figure}

\subsection{Main Results}
\textbf{Performance Comparison with SOTA Methods.}
In Table~\ref{tab:sota}, the performance of the proposed \Ours is compared with the state-of-the-art (SOTA) models. 
On CUB-200-2011 dataset, \Ours achieves localization accuracy of Top-1 87.0\%, Top-5 96.1\%, which surpasses the SOTA methods by significant margins. Specifically, \Ours achieves surprisingly 98.0\% localization accuracy under GT-known metric, which shows the effectiveness of introducing generative model for WSOL. \Ours outperforms the SOTA method $\text{C}^{2}\text{AM}$~\cite{DBLP:conf/cvpr/XieXCHZS22} by 5.2\% (87.0\% $vs.$ 81.8\%) and 5.0\% (96.1\% $vs.$ 91.1\%) under Top-1 Loc and Top-5 Loc metrics respectively.
When solely considering the localization performance (using GT-known Loc metric), \Ours significantly outperforms the SOTA method SCM~\cite{bai2022weakly} by 1.4\% (98.0\% $vs.$ 96.6\%).
On the more challenging ImageNet-1K dataset, \Ours also significantly outperforms the SOTA method $\text{C}^{2}\text{AM}$~\cite{DBLP:conf/cvpr/XieXCHZS22} and $\text{BAS}$~\cite{DBLP:conf/cvpr/WuZC22} by 5.6\% (65.2\% $vs.$ 59.6\%), 6.0\% (73.4\% $vs.$ 67.4\%), and 3.2\% (75.0\% $vs.$ 71.8\%) under Top-1 Loc, Top-5 Loc and GT-known Loc metrics respectively.
Such strong results clearly demonstrate the superiority of the generative model over conventional discriminative models for weakly supervised object localization.

\textbf{Localization Results with respect to Prompt Embeddings.}
The localization results of \Ours are shown in Fig.~\ref{fig:viz_query}.
In the first row of Fig.~\ref{fig:viz_query}(upper), for the image with category \texttt{spaniel}, the discriminative prompt embedding $f_d$ fails to activate the legs, while the representative prompt embedding $f_r$ activates full object extent but suffering from the background noise. By combining $f_d$ and $f_r$, $f_c$ fully activates the object regions while maintaining low background noise. 
In the second row of Fig.~\ref{fig:viz_query}(upper), we show the localization maps with prompt embeddings generated by different categories. 
Interestingly, categories that are highly related to \texttt{spaniel} ($i.e.,$ \texttt{dog, husky}) can also correctly activate the foreground object, which indicates that \Ours is robust to the classifiers.
When using the categories that are less related to \texttt{spaniel} ($i.e.$, \texttt{man, tower}) will introduce too much background noise and fail to localize the objects. 

As shown in Fig.~\ref{fig:viz_query}(lower), for the test images which contain multiple objects from various classes, \Ours is able to generate high quality localization maps when given the corresponding prompt embeddings (generated using corresponding categories). The result demonstrates \Ours can not only generate representative localization maps but also be able to discriminate object categories, revealing the potential of extending \Ours to more challenging weakly supervised object detection or segmentation task.

\textbf{Statistical Result with respect to Token Embeddings.}
In Fig.~\ref{fig:vrvd}, the embedding vectors of the \textit{meta} token are uniformly distributed in the two-dimensional feature space (yellow dots). After learning the representative embeddings/features of the categories, the embedding vectors of the \textit{concept} token become uneven (blue dots), $i.e.$ less discriminative indicated by less deviation $\sigma_x$ and $\sigma_y$, 
which reveals the inconsistency between representative and discriminative embeddings.

{\color{blue}
%QXYE: 这里的V_r, V_d似乎没有很好的定义和说明？ 正文中说的也不清晰
\begin{figure}[t]
	\includegraphics[width=1\linewidth]{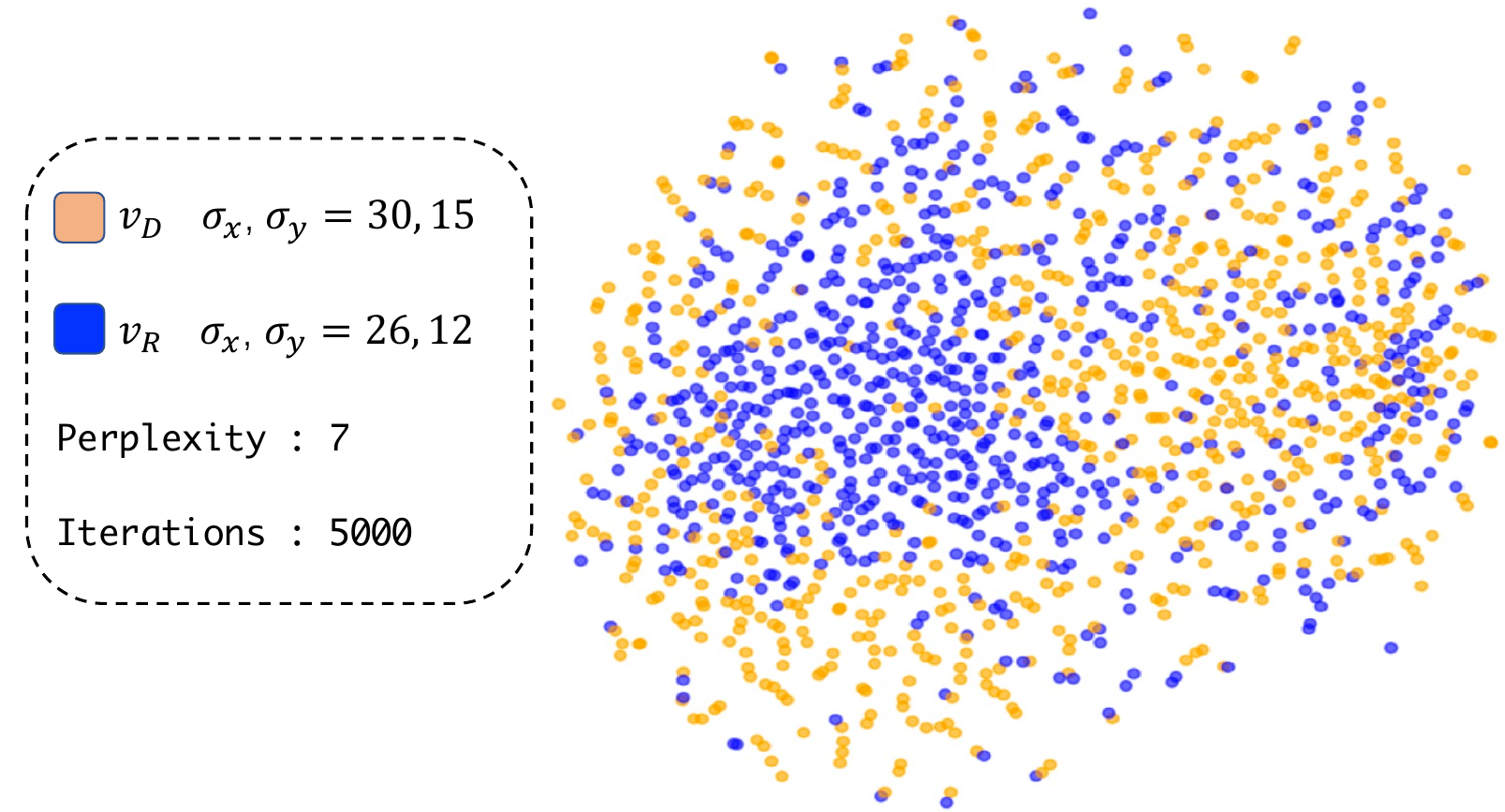}
	\caption{{Statistical result of the embedding vectors of the \textit{meta} token and the \textit{concept} token for ImageNet-1K categories using tSNE.} $\sigma_x,\sigma_y$ respectively denote the standard deviation of the points in the $x$ and $y$ directions.}
\label{fig:vrvd}
\end{figure}
}

\begin{table}[t]
    \centering
    \small
    \setlength{\tabcolsep}{1mm}
    
\begin{tabular}{c|clccc}
\toprule
& \multirow{2}{*}{Finetune} & \multirow{2}{*}{Embedding} & \multicolumn{3}{c}{ImageNet-1K}\tabularnewline
\cline{4-6}
 & &  & Top-1 Loc & Top-5 Loc & GT-kno. Loc\tabularnewline
\midrule 
1 & & $f_d$ & 61.2 & 69.0 & 70.4\tabularnewline
% \hline
2 & & $f_r$(w/o init) & 44.6 & 50.2 & 51.3\tabularnewline
3 & & $f_r$ & 64.0 & 72.1 & 73.7\tabularnewline
4 & & $f_c$(w/o init) & 56.2 & 63.2 & 64.5\tabularnewline
5 & & $f_c$ & 64.5 & 72.7 & 74.2\tabularnewline
% \hline 
% rerun
\midrule
6 & \checkmark & $f_d$ & 62.0 & 69.8 & 71.4\tabularnewline
7 &\checkmark & $f_r$ & 64.9 & 73.1 & 74.6\tabularnewline
8 & \checkmark & $f_c$ & \textbf{65.1} & \textbf{73.3} & \textbf{74.9}\tabularnewline
\bottomrule
\end{tabular}

    \caption{{Ablation studies} of \Ours.}
    \label{tab:ablation_tok}
\end{table}

\subsection{Ablation Study}

\textbf{Baseline.} We build the baseline method by using solely the discriminative embedding $f_d$ (Line 1 of Table~\ref{tab:ablation_tok}). The performance (61.2\% Top-1 Loc Acc.) of the baseline outperforms the SOTA methods, which indicates the great advantage of introducing the generative model for WSOL.

\textbf{Representative Embedding.} When using the trained representative embedding $f_r$ initialized by $f_d$ (Line 3 of Table~\ref{tab:ablation_tok}), \Ours outperforms the baseline by 2.8\% (64.0\% $vs.$ 61.2\%) under Top-1 Loc metric, which validates the importance of the representative features of categories for object localization. We also train representative embedding $f_r$ which is random initialized (Line 2 of Table~\ref{tab:ablation_tok}). The performance of $f_r$ drops to very low-level suffering from the local optimal embeddings.

\begin{table*}[t]
    \centering
    \small 
    \setlength{\tabcolsep}{1mm}

\begin{tabular}{ r cccccccc}
\toprule
\multirow{2}{*}{Method} & \multirow{2}{*}{Loc Back.} & \multirow{2}{*}{Cls Back.} & \multirow{2}{*}{Params.} &\multicolumn{5}{c}{ImageNet-1K} \tabularnewline
\cline{5-9}
 &   &   & &Top-1 Loc & Top-5 Loc & GT-known Loc & Top-1 Cls & Top-5 Cls\tabularnewline
\midrule

TS-CAM~\cite{DBLP:conf/iccv/GaoWPP0HZY21} & \multicolumn{2}{c}{Deit-S (ImageNet-1K)} & 22.4M & 53.4 & 64.3 & 67.6 & 74.3 & 92.1\tabularnewline
TS-CAM~\cite{DBLP:conf/iccv/GaoWPP0HZY21} & \multicolumn{2}{c}{ViT-H (LAION-2B~\cite{DBLP:journals/corr/abs-2210-08402}, ImageNet-1K)} & 633M & 42.1 & 49.9 & 52.2 & 77.4 & 93.7\tabularnewline
\hline \rowcolor{gray0}
\rowcolor{gray0} GenPromp$\dagger$ & Stable Diffusion & EfficientNet-B7 & 1017M + 66M & \textbf{65.2} & \textbf{73.4} & \textbf{75.0} & 85.1 & 97.2\tabularnewline
\bottomrule
\end{tabular}

    \caption{Performance comparison with respect to model size and training data. With a larger backbone and pre-training dataset, the discriminatively trained method TS-CAM does not achieve higher performance.}
    \vspace{-0.2cm}
    \label{tab:ablation_param}
\end{table*}

\begin{figure}[tbp]
	\includegraphics[width=0.99\linewidth]{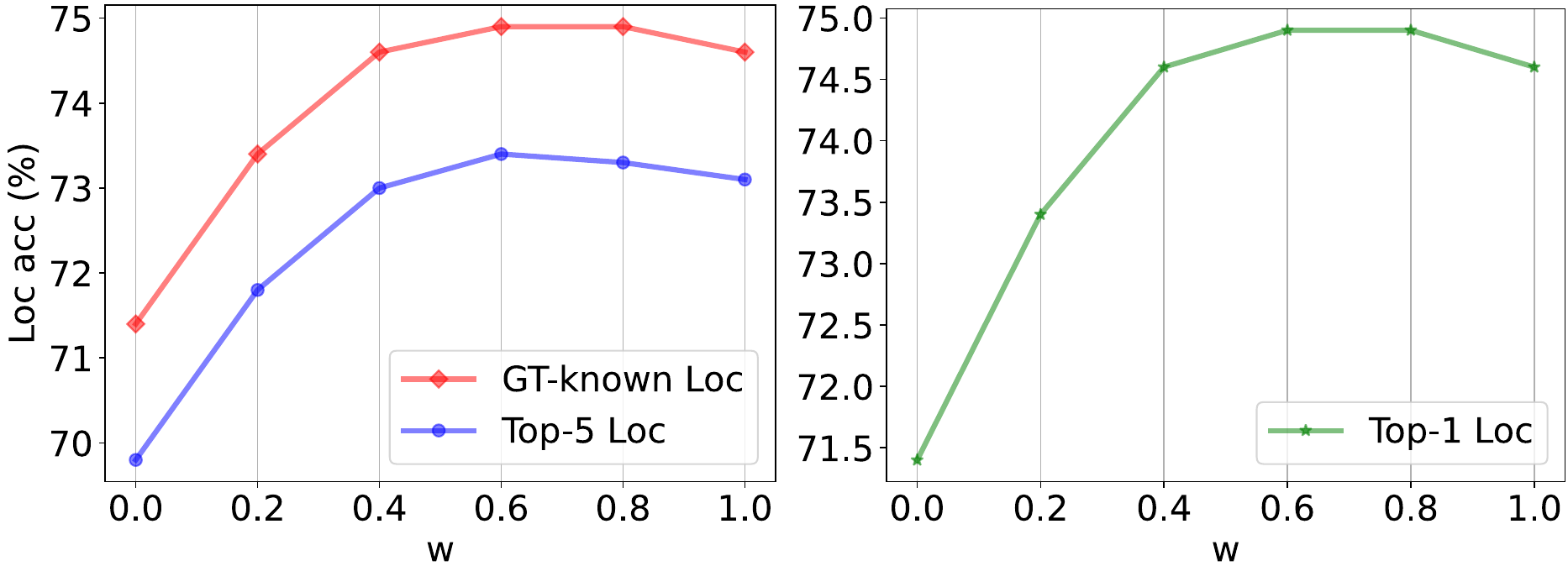}
	\caption{Localization accuracy under IoUs on ImageNet-1K using different embedding combination weights $w$.}
\label{fig:weight}
\end{figure}

\textbf{Embedding Combination.} By combining $f_r$ and $f_d$, the performance of \Ours can be further boost to 64.5\% in Top-1 Loc (Line 5 of Table~\ref{tab:ablation_tok}). We also conduct experiments to show the effect of the combination weight $w$ on ImageNet 1K dataset. The performance keeps consistent when the combination weight $w$ falls to $[0.4,0.8]$, as shown in Fig.~\ref{fig:weight}. With a proper $w$, \Ours is able to generate discriminative while representative prompt embeddings and active full object extent with the least background noise.

\textbf{Model Finetuning.} By filling the domain gap through finetuning the backbone network (attention-unet), a performance gain of $0.6\%$ (65.1\% $vs.$ 64.5\%) can be achieved in Top-1 Loc (Line 8 of Table~\ref{tab:ablation_tok}).

\begin{table}[t]
    \centering
    % \footnotesize 
    % \renewcommand\arraystretch{1.0}
    \small
    \setlength{\tabcolsep}{1mm}
    \begin{tabular}{ r ccc}
\toprule
\multirow{2}{*}{Timesteps} & \multicolumn{3}{c}{ImageNet-1K}\tabularnewline
\cline{2-4}
& Top-1 Loc & Top-5 Loc & GT-known Loc\tabularnewline
\midrule 
% \hline 
1 & 64.6 & 72.7 & 74.3\tabularnewline
% \hline 
10 & 64.8 & 72.9 & 74.5\tabularnewline
100 & 64.9 & 73.1 & 74.6\tabularnewline
200 & 64.6 & 72.7 & 74.3\tabularnewline
500 & 59.4 & 66.9 & 68.3\tabularnewline
1000 & 42.8 & 48.3 & 49.3\tabularnewline
1+100 & \textbf{65.1} & \textbf{73.3} & \textbf{74.9}\tabularnewline
% rerun
1+100+200 & \textbf{65.1} & \textbf{73.3} & \textbf{74.9}\tabularnewline
1+100+200+500 & 64.1 & 72.2 & 73.7\tabularnewline
\bottomrule
\end{tabular}
    \vspace{-0.1cm}
    \caption{{Performance under test timesteps.}}
    % \vspace{+2mm}
    
    % \vspace{+2mm
    \label{tab:ablation_timesteps}
    % \vspace{-2mm}
\end{table}

\textbf{Model Size and Training Data.} In Table~\ref{tab:ablation_param}, we re-implement TS-CAM with larger backbone (\eg{ViT-H}) and more training data (\eg{LAION-2B}). The re-implemented TS-CAM achieves higher classification accuracy while much lower localization accuracy compared to the Deit-S based TS-CAM. We attribute this to the inherent flaw of the discriminatively trained classification model, $i.e.$, local discriminative regions are capable of minimizing image classification loss, but experience difficulty in accurate object localization. A larger backbone and more training data even make this phenomenon even more serious.

\begin{table}[t]
    \centering
    \small
    \setlength{\tabcolsep}{1mm}
    \begin{tabular}{ r ccc}
\toprule
\multirow{2}{*}{Resolution} & \multicolumn{3}{c}{ImageNet-1K}\tabularnewline
\cline{2-4}
& Top-1 Loc & Top-5 Loc & GT-known Loc\tabularnewline
\midrule 
% \hline 
8 & 58.5 & 65.7 & 67.1\tabularnewline
% \hline 
16 & 62.5 & 70.5 & 72.0\tabularnewline
32 & 38.0 & 42.7 & 43.6\tabularnewline
64 & 17.5 & 19.9 & 20.3\tabularnewline
% rerun
8+16 & \textbf{65.1} & \textbf{73.3} & \textbf{74.9} \tabularnewline
8+16+32 & 64.2 & 72.2 & 73.7\tabularnewline
\bottomrule
\end{tabular}

    \vspace{-0.1cm}
    \caption{{Performance under attention map resolutions.}}
    \label{tab:ablation_resolution}
\end{table}
\textbf{Effect of Timesteps and Resolutions.} In Tables~\ref{tab:ablation_timesteps} and ~\ref{tab:ablation_resolution}, we evaluate the performance by aggregating attention maps under different timesteps and spatial resolutions. It can be seen that aggregating the attention maps of spatial resolutions $8\times 8$ and $16\times 16$ at time steps $1$, $100$ produces the best localization performance.

\section{Conclusion and Future Remark} 
To solve the partial object activation brought by discriminatively trained activation models in a systematic way, we propose \Ours, which formulates WSOL as a conditional image denoising procedure. During training, we use the learnable embeddings to conditionally recover the input image with the aim to learn the representative embeddings. During inference, \Ours first combines the trained representative embeddings with another discriminative embeddings from the pre-trained CLIP model. 
The combined embeddings are then used to generate attention maps at multiple timesteps and resolutions, which facilitate localizing the full object extent.
\Ours not only sets a solid baseline for WSOL with generative models but also provides fresh insight for handling vision tasks using vision-language models.

When claiming the advantages of \Ours, we also realize its disadvantages. One major disadvantage is its dependency on large-scale pre-trained vision-language models, which could slow down the inference speed and raise the requirement for GPU memory cost. Such a disadvantage should be solved in future work.

{\small
\bibliographystyle{ieee_fullname}
\bibliography{arxiv}
}

\appendix

\begin{table}[tbp]
    \centering
    \footnotesize
    \setlength{\tabcolsep}{1mm}

\tablestyle{4pt}{1.2}
\begin{tabular}{l|c|r|l|c}
\toprule
Dataset & Ann. & \# Images & How to collect & $t$ (s/img)\tabularnewline
\midrule
CUB-200-2011~\cite{WahCUB_200_2011} & $\mathcal{I}$ &11,788 & Manual & 1.5\tabularnewline
Imagenet~\cite{DBLP:journals/ijcv/RussakovskyDSKS15} & $\mathcal{I}$ & 14,197,122 & Manual & 1.5\tabularnewline
JFT-3B~\cite{DBLP:conf/iclr/DosovitskiyB0WZ21} & $\mathcal{I}^{\dagger}$ & 3,000,000,000 & Semi-automatic & $\approx0$\tabularnewline
\hline
CC12M~\cite{DBLP:conf/cvpr/ChangpinyoSDS21} & $\mathcal{T}^{\dagger}$ & 12,000,000 & Web crawler & $\approx0$ \tabularnewline
WIT~\cite{DBLP:conf/icml/RadfordKHRGASAM21} & $\mathcal{T}^{\dagger}$ &400,000,000 & Web crawler & $\approx0$\tabularnewline
LAION-400M~\cite{DBLP:journals/corr/abs-2111-02114} & $\mathcal{T}^{\dagger}$ & 400,000,000 & Web crawler & $\approx0$\tabularnewline
LAION-5B~\cite{DBLP:journals/corr/abs-2210-08402} & $\mathcal{T}^{\dagger}$ & 5,850,000,000 & Web crawler & $\approx0$\tabularnewline
\hline
Cityscapes~\cite{DBLP:conf/cvpr/CordtsORREBFRS16} & $\mathcal{B}$ & 25,000 & Manual & 37.5\tabularnewline
COCO~\cite{DBLP:conf/eccv/LinMBHPRDZ14} & $\mathcal{B}$ & 328,000 & Manual & 37.5\tabularnewline
\bottomrule
\end{tabular}

    \vspace{-0.1cm}
    \caption{\textbf{The size and data collecting approaches of some commonly used datasets}. $\mathcal{I}, \mathcal{T}, \mathcal{B}$ denotes the image category labels, text descriptions and bounding box annotations respectively. $t$ denotes the average annotation time per image. $\dagger$ indicates the annotation is noisy. }
       
    \label{tab:annotation_cost}
    % \vspace{-2mm}
\end{table}

\section{Annotation Cost}
As shown in Table~\ref{tab:annotation_cost}, we compare the size and data collection methods of commonly used datasets with three types of annotations: image category labels, text descriptions, and bounding boxes.
It can be observed that datasets with accurate bounding box labels, such as Cityscapes and COCO, are usually small in size due to the high cost of manual annotation. However, when using image category labels, the dataset size can be increased to 14 million (e.g., ImageNet).
For datasets with a huge size, such as JFT-3B, WIT, and LAION-5B, manual annotation becomes impractical. Instead, semi-automatic annotation methods or web crawler algorithms are used to extensively collect noisy annotated data.
Thanks to the rapid development of the Internet, a large number of image-text pairs can be found in websites, forums, and libraries, which are naturally annotated by citizens and can be easily obtained by crawler algorithms. Since collecting image-text pairs hardly requires human participation, their annotation cost is negligible.
In this paper, the proposed method \Ours is implemented based the Stable Diffusion model, which is pre-trained on LAION-5B. Accordingly, \Ours hardly introduces additional annotation cost for a weakly supervised learning system.

\begin{figure}[t]
% 	\centering
	\includegraphics[width=0.99\linewidth]{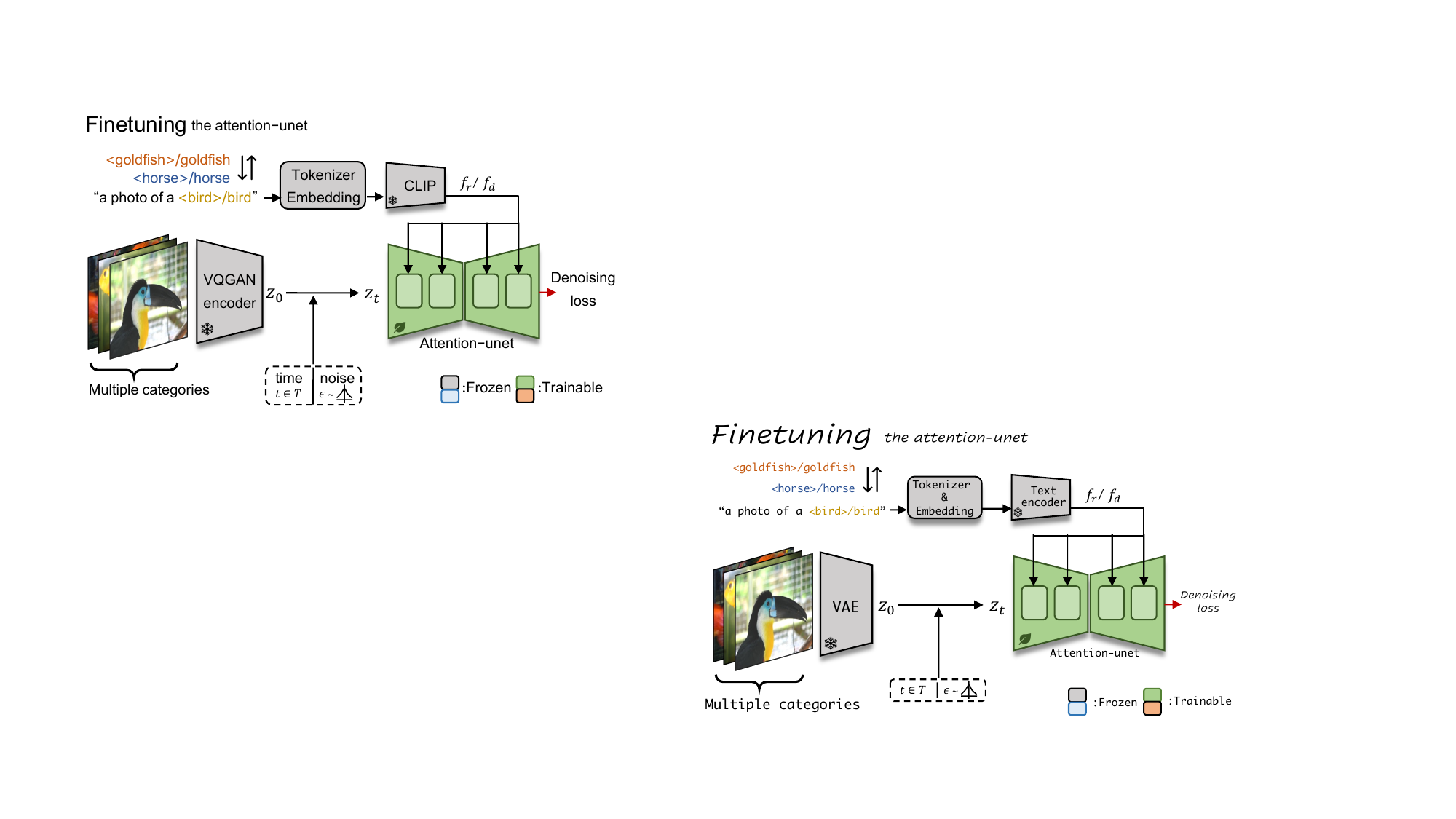}
	% \vspace{-0.2cm}
	\caption{\textbf{Finetuning pipeline of \Ours.} }
	% \vspace{-0.2cm}
\label{fig:finetune}
\end{figure}

\section{Finetuning}
As described in Section 4 in the main document, after obtaining the representative embeddings $f_r^{*}$ for all image categories, we finetune the attention-unet (parameterized by $\epsilon_{\theta}$) to reduce the domain gap between the model and the target dataset. We demonstrate the pipeline of finetuning in Fig.~\ref{fig:finetune}. The finetuning pipeline is different from the training pipeline (shown in Fig. 3 of the main document) in two aspects: (1) Each training batch contains images of multiple categories, (2) The prompt embeddings are frozen while the attention-unet is trainable.

\begin{figure*}[htbp]
% 	\centering
	\includegraphics[width=0.99\linewidth]{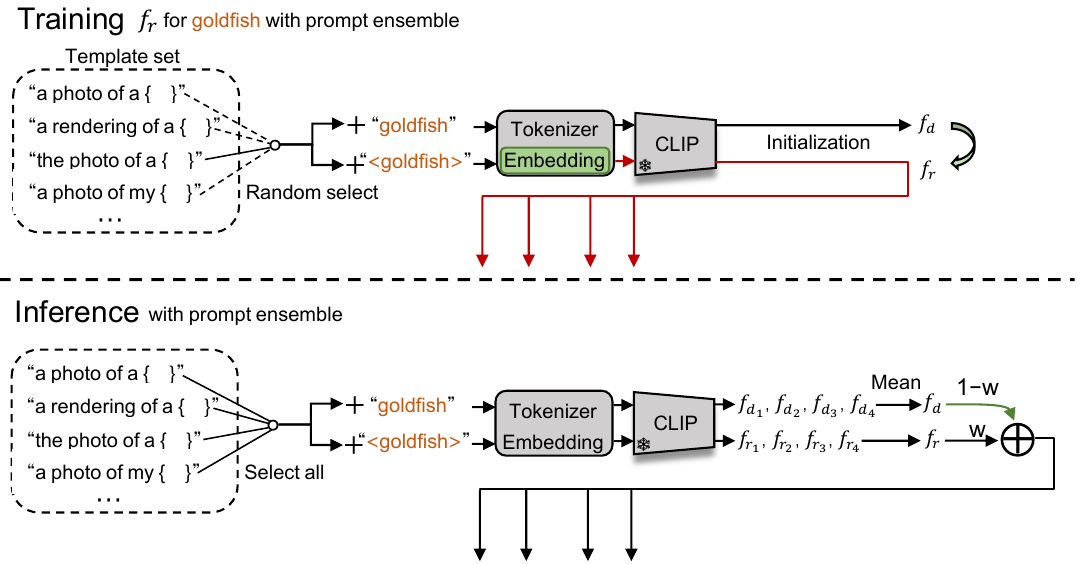}
	\caption{\textbf{Workflow of the proposed prompt ensemble strategy}. The image encoding and activate map generation procedure are omitted for clarity.}
\label{fig:prompt_ensemble}
\end{figure*}

\section{Prompt Ensemble}
To further improve the performance of \Ours, we propose a prompt ensemble strategy.
As shown in Fig.~\ref{fig:prompt_ensemble}, during training, we random select a template from a template set. Then, we respectively fill the $\textit{meta}$ token (\texttt{goldfish}) and the $\textit{concept}$ token ($\langle\texttt{goldfish}\rangle$) into the template to obtain the two input prompts, which are used to learn the representative embedding $f_r$. 
During inference, for each template in the template set, we combines it with the two tokens to form the input prompts. Then, all the prompts are encoded into prompt embeddings by the pre-trained CLIP model. After that, the discriminative embedding $f_d$ is obtained by averaging the discriminative embeddings generated by different templates (\eg{$f_{d_1},f_{d_2},f_{d_3},f_{d_4}$ in Fig.~\ref{fig:prompt_ensemble}), the representative embedding $f_r$ is obtained by averaging the representative embeddings generated by different templates (\eg{$f_{r_1},f_{r_2},f_{r_3},f_{r_4}$ in Fig.~\ref{fig:prompt_ensemble}). Finally, $f_d$ and $f_r$ are combined to $f_c$, which is fed into the network to generate attention maps. In experiments, we use a template set that consists of 7 templates:

``\texttt{a photo of a \{ \}}''

``\texttt{a rendering of a \{ \}}''

``\texttt{the photo of a \{ \}}''

``\texttt{a photo of my \{ \}}''

``\texttt{a photo of the \{ \}}''

``\texttt{a photo of one \{ \}}''

``\texttt{a rendition of a \{ \}}''

\begin{table}[tbp]
    \centering
    \footnotesize
    \setlength{\tabcolsep}{1mm}
    \begin{tabular}{ccccccc}
\toprule
\multirow{2}{*}{Embedding} & \multicolumn{6}{c}{ImageNet-1K}\tabularnewline
\cline{2-7}
& Top-1 Loc & Top-5 Loc & GT-known Loc & M-Ins & Part & More\tabularnewline
\midrule 
% \hline 
$f_d$ & 62.2 & 70.0 & 71.5 & 9.4 & 3.8 & 8.2\tabularnewline
$f_r$ & 64.9 & 73.1 & 74.7 & 9.5 & 2.4 & 7.3\tabularnewline
$f_c$ & 65.2 & 73.4 & 75.0 & 9.1 & 3.0 & 6.9\tabularnewline

\bottomrule
\end{tabular}

    \vspace{-0.1cm}
    \caption{\textbf{Localization error statistics.} The results are correspond to row 17-19 in Table~\ref{tab:ablation}. ``M-Ins'', ``Part'' and ``More'' denote the multi-instance error, localization part error and localization more error respectively.}
    \label{tab:ablation_part_more}
\end{table}

\begin{table}[tbp]
    \centering
    \small
    \setlength{\tabcolsep}{1mm}
    
\begin{tabular}{cccc}
\toprule
\multirow{2}{*}{Noise $\epsilon$} & \multicolumn{3}{c}{ImageNet-1K}\tabularnewline
\cline{2-4}
& Top-1 Loc & Top-5 Loc & GT-known Loc\tabularnewline
\midrule 
% \hline 
 & 64.8 & 73.0 & 74.6\tabularnewline
\checkmark & \textbf{65.1} & \textbf{73.3} & \textbf{74.9}\tabularnewline
\bottomrule
\end{tabular}

    \vspace{-0.1cm}
    \caption{\textbf{Evaluation of noise levels in the inference time.} For the experiment that includes noise $\epsilon$ in the inference time, we conduct 10 experiments under different random seeds and average their results as the final result.}
    \label{tab:ablation_noise}
\end{table}

\begin{table*}[htbp]
    \centering
    \footnotesize
    \setlength{\tabcolsep}{1.2mm}
    \begin{tabular}{ r cccccccc}
\toprule
\multirow{2}{*}{Method} & \multirow{2}{*}{Loc Back.} & \multirow{2}{*}{Cls Back.} & \multicolumn{3}{c}{CUB-200-2011} & \multicolumn{3}{c}{ImageNet-1K}\tabularnewline
\cline{4-9}
 &  &  & Top-1 Loc & Top-5 Loc & GT-known Loc & Top-1 Loc & Top-5 Loc & GT-known Loc\tabularnewline
\midrule
CAM$_{\text{CVPR'16}}$~\cite{DBLP:conf/cvpr/ZhouKLOT16} & \multicolumn{2}{c}{VGG16} & 41.1 & 50.7 & 55.1 & 42.8 & 54.9 & 59.0\tabularnewline
ADL$_{\text{CVPR'19}}$~\cite{DBLP:journals/pami/ChoeLS21} & \multicolumn{2}{c}{VGG16} & 52.4 & - & 75.4 & 44.9 & - & -\tabularnewline
DANet$_{\text{ICCV'19}}$~\cite{DBLP:conf/iccv/XueLWJJY19} & \multicolumn{2}{c}{VGG16} & 52.5 & 62.0 & 67.7 & - & - & -\tabularnewline
SLT$_{\text{CVPR'21}}$~\cite{DBLP:conf/cvpr/GuoHWZ21} & \multicolumn{2}{c}{VGG16} & 67.8 & - & 87.6 & 51.2 & 62.4 & 67.2\tabularnewline
FAM$_{\text{ICCV'21}}$~\cite{DBLP:conf/iccv/MengZ00021} & \multicolumn{2}{c}{VGG16} & 69.3 & - & 89.3 & 52.0 & - & 71.7\tabularnewline
TAFormer$_{\text{TPAMI'22}}$~\cite{9996553} & \multicolumn{2}{c}{VGG16} & 72.0 & 85.9 & 90.8 & 53.4 & 67.7 & 74.0\tabularnewline
BAS$_{\text{CVPR'22}}$~\cite{DBLP:conf/cvpr/WuZC22} & \multicolumn{2}{c}{VGG16} & 71.3 & 85.3 & 91.1 & 53.0 & 65.4 & 69.6\tabularnewline
\hline
CAM$_{\text{CVPR'16}}$~\cite{DBLP:conf/cvpr/ZhouKLOT16} & \multicolumn{2}{c}{MobileNetV1} & 48.1 & 59.2 & 63.3 & 43.4 & 54.4 & 59.0\tabularnewline
HaS$_{\text{ICCV'17}}$~\cite{DBLP:conf/iccv/SinghL17} & \multicolumn{2}{c}{MobileNetV1} & 46.7 & - & 67.3 & 42.7 & - & 60.1\tabularnewline
ADL$_{\text{CVPR'19}}$~\cite{DBLP:journals/pami/ChoeLS21} & \multicolumn{2}{c}{MobileNetV1} & 47.7 & - & - & 43.0 & - & -\tabularnewline
FAM$_{\text{ICCV'21}}$~\cite{DBLP:conf/iccv/MengZ00021} & \multicolumn{2}{c}{MobileNetV1} & 65.7 & - & 85.7 & 46.2 & - & 62.1\tabularnewline
TAFormer$_{\text{TPAMI'22}}$~\cite{9996553} & \multicolumn{2}{c}{MobileNetV1} & 66.7 & 80.2 & 85.0 & 47.6 & 65.5 & 68.8\tabularnewline
BAS$_{\text{CVPR'22}}$~\cite{DBLP:conf/cvpr/WuZC22} & \multicolumn{2}{c}{MobileNetV1} & 69.8 & 86.0 & 92.4 & 53.0 & 66.6 & 72.0\tabularnewline
\hline
CAM$_{\text{CVPR'16}}$~\cite{DBLP:conf/cvpr/ZhouKLOT16} & \multicolumn{2}{c}{ResNet50} & 46.7 & 54.4 & 57.4 & 39.0 & 49.5 & 51.9\tabularnewline
ADL$_{\text{CVPR'19}}$~\cite{DBLP:journals/pami/ChoeLS21} & \multicolumn{2}{c}{ResNet50-SE} & 62.3 & - & - & - & - & 48.5\tabularnewline
FAM$_{\text{ICCV'21}}$~\cite{DBLP:conf/iccv/MengZ00021} & \multicolumn{2}{c}{ResNet50} & 73.7 & - & 85.7 & 54.5 & - & 64.6\tabularnewline
SPOL$_{\text{CVPR'21}}$~\cite{DBLP:conf/cvpr/WeiWLWZC21} & \multicolumn{2}{c}{ResNet50} & 80.1 & 93.4 & 96.5 & 59.1 & 67.2 & 69.0\tabularnewline
TAFormer$_{\text{TPAMI'22}}$~\cite{9996553} & \multicolumn{2}{c}{ResNet50} & 75.0 & 87.8 & 91.2 & 57.5 & 69.9 & 75.5\tabularnewline
DA$_{\text{CVPR'22}}$~\cite{DBLP:conf/cvpr/ZhuSCYWL22} & \multicolumn{2}{c}{ResNet50} & 66.7 & - & 81.8 & 55.8 & - & 70.3\tabularnewline
BAS$_{\text{CVPR'22}}$~\cite{DBLP:conf/cvpr/WuZC22} & \multicolumn{2}{c}{ResNet50} & 77.3 & 90.1 & 95.1 & 57.2 & 67.4 & 71.8\tabularnewline
\hline
CAM$_{\text{CVPR'16}}$~\cite{DBLP:conf/cvpr/ZhouKLOT16} & \multicolumn{2}{c}{InceptionV3} & 41.1 & 50.7 & 55.1 & 46.3 & 58.2 & 62.7\tabularnewline
DANet$_{\text{ICCV'19}}$~\cite{DBLP:conf/iccv/XueLWJJY19} & \multicolumn{2}{c}{InceptionV3} & 49.5 & 60.5 & 67.0 & 47.5 & 58.3 & -\tabularnewline
SLT$_{\text{CVPR'21}}$~\cite{DBLP:conf/cvpr/GuoHWZ21} & \multicolumn{2}{c}{InceptionV3} & 66.1 & - & 86.5 & 55.7 & 65.4 & 67.6\tabularnewline
FAM$_{\text{ICCV'21}}$~\cite{DBLP:conf/iccv/MengZ00021} & \multicolumn{2}{c}{InceptionV3} & 70.7 & - & 87.3 & 55.2 & - & 68.6\tabularnewline
TAFormer$_{\text{TPAMI'22}}$~\cite{9996553} & \multicolumn{2}{c}{InceptionV3} & 73.3 & 84.1 & 88.7 & 56.0 & 66.5 & 69.8\tabularnewline
BAS$_{\text{CVPR'22}}$~\cite{DBLP:conf/cvpr/WuZC22} & \multicolumn{2}{c}{InceptionV3} & 73.3 & 86.3 & 92.2 & 58.5 & 69.0 & 71.9\tabularnewline
CREAM$_{\text{CVPR'22}}$~\cite{DBLP:conf/cvpr/XuHZFZZLG22} & \multicolumn{2}{c}{InceptionV3} & 71.8 & 86.4 & 90.4 & 56.1 & 66.2 & 69.0\tabularnewline
\hline
TS-CAM$_{\text{ICCV'21}}$~\cite{DBLP:conf/iccv/GaoWPP0HZY21} & \multicolumn{2}{c}{Deit-S} & 71.3 & 83.8 & 87.7 & 53.4 & 64.3 & 67.6\tabularnewline
LCTR$_{\text{AAAI'22}}$~\cite{DBLP:conf/aaai/ChenWWJSTWZC22} & \multicolumn{2}{c}{Deit-S} & 79.2 & 89.9 & 92.4 & 56.1 & 65.8 & 68.7\tabularnewline
SCM$_{\text{ECCV'22}}$~\cite{bai2022weakly} & \multicolumn{2}{c}{Deit-S} & 76.4 & 91.6 & 96.6 & 56.1 & 66.4 & 68.8\tabularnewline
\hline
 % &  &  &  &  &  &  &  & \tabularnewline
%
DiPS$_{\text{WACVW'23}}$~\cite{DBLP:conf/wacv/MurtazaBPSG23} & Deit-S & TransFG~\cite{he2022transfg} & 88.2 & - & - & - & - & -\tabularnewline
PSOL$_{\text{CVPR'20}}$~\cite{DBLP:conf/cvpr/ZhangCW20} & DenseNet161 & EfficientNet-B7 & 80.9 & 90.0 & 91.8 & 58.0 & 65.0 & 66.3\tabularnewline
C$^2$AM$_{\text{CVPR'22}}$~\cite{DBLP:conf/cvpr/XieXCHZS22} & DenseNet161 & EfficientNet-B7 & 81.8 & 91.1 & 92.9 & 59.6 & 67.1 & 68.5 \tabularnewline
\hline \rowcolor{gray0}
\rowcolor{gray0} GenPromp (Ours) & Stable Diffusion & EfficientNet-B7 & 87.0 & 96.1 & \textbf{98.0} & 65.1 & 73.3 & 74.9 \tabularnewline
\rowcolor{gray0} GenPromp$\dagger$ (Ours) & Stable Diffusion & EfficientNet-B7 & 87.0 & 96.1 & \textbf{98.0} & \textbf{65.2} & \textbf{73.4} & \textbf{75.0} \tabularnewline
\rowcolor{gray0} GenPromp$\dagger$ (Ours) & Stable Diffusion & TransFG~\cite{he2022transfg} & \textbf{89.3} & \textbf{96.5} & \textbf{98.0} & - & - & - \tabularnewline
\bottomrule
\end{tabular}

    \caption{\textbf{Performance comparison} of the proposed \Ours approach with the state-of-the-art methods on the CUB-200-2011 test set and ImageNet-1K validation set. \textit{Loc Back}. denotes the localization backbone, \textit{Cls Back}. the backbone for classification, and $\dagger$ the prompt ensemble strategy, which ensembles the localization results from multiple prompts.}
    \label{tab:sota2}
\end{table*}

\section{Additional Experimental Results}

\textbf{Complete Performance Comparison with SOTA Methods.}
Table~\ref{tab:sota2} shows the complete performance comparison of the proposed \Ours and the state-of-the-art (SOTA) models (extension of Table 1 in the main document). On CUB-200-2011 and ImageNet-1K dataset, \Ours surpasses the SOTA methods by significant margins.
Such strong results clearly demonstrate the superiority of the generative model over conventional discriminative models for weakly supervised object localization.

\textbf{Localization Error Analysis.} To further reveal the effect of the proposed prompt embeddings  (\eg{$f_d$, $f_r$, $f_c$), following TS-CAM~\cite{DBLP:conf/iccv/GaoWPP0HZY21}, we evaluate the localization errors of: multi-instance error (M-Ins), localization part error (Part), and localization more error (More). They are respectively defined as follows.
\begin{itemize}
    \item M-Ins indicates that the predicted bounding box intersects with at least two ground-truth boxes, and $\text{IoG} > 0.3$.
    \item Part indicates that the predicted bounding box only cover the parts of object, and $\text{IoP} > 0.5$. 
    \item More indicates that the predicted bounding box is larger than the ground truth bounding box by a large margin, and $\text{IoG} > 0.7$. 
\end{itemize}
where IoG and IoP are defined as Intersection over Ground truth box and Intersection over Predict bounding box, respectively (similar to IoU (Intersection over Union)).
Each metric calculates the percentage of images belonging to corresponding error in the validation/test set. 
Please refer to TS-CAM~\cite{DBLP:conf/iccv/GaoWPP0HZY21} for a detailed definition of the three metrics.
Table~\ref{tab:ablation_part_more} lists localization error statistics of M-Ins, Part, and More. 
Compare to the discriminative embedding $f_d$, the learned representative embedding $f_r$ reduces both Part and More errors by 1.4\% (3.8\% $vs.$ 2.4\%) and 0.9\% (8.2\% $vs.$ 7.3\%) respectively, demonstrating that the representative embedding alleviates the partial object activation problem. By combining the representative embedding $f_r$ with the discriminative embedding $f_d$, the More errors drop 0.4\% (7.3\% $vs.$ 6.9\%) while the Part errors increase 0.6\% (3.0\% $vs.$ 2.4\%) compared to $f_r$. This demonstrates that $f_c$ can further depress the background noise while keeping relatively low Part errors.

\textbf{Effect of Noise $\epsilon$.} In Table~\ref{tab:ablation_noise}, we evaluate the performance by setting the noise $\epsilon$ (in Eq. 1 and Eq. 2 of the main document) to 0 during inference. Without noise $\epsilon$, the performance of \Ours drops 0.3\% in Top-1 Loc in average. Similar to the methods~\cite{DBLP:journals/pami/ChoeLS21,DBLP:conf/cvpr/ChoeS19,DBLP:conf/cvpr/MaiYL20,DBLP:conf/iccv/SinghL17,DBLP:conf/iccv/YunHCOYC19,DBLP:conf/cvpr/ZhangWF0H18} based on adversarial erasing, the input noise in \Ours can also alleviate the part activation issue, which drives the network to mine the representative yet less discriminative object parts.

\begin{table*}
    % \caption{Global caption}
    \begin{minipage}{1.\linewidth}
        \centering
        \footnotesize
         \setlength{\tabcolsep}{1mm}
    \begin{tabular}{ r cccccccc}
\toprule
\multirow{2}{*}{Method} & \multirow{2}{*}{Loc Back.} & \multirow{2}{*}{Cls Back.} & \multirow{2}{*}{Params.} &\multicolumn{5}{c}{CUB-200-2011} \tabularnewline
\cline{5-9}
 &   &   & &Top-1 Loc & Top-5 Loc & GT-known Loc & Top-1 Cls & Top-5 Cls\tabularnewline
\midrule

TS-CAM~\cite{DBLP:conf/iccv/GaoWPP0HZY21} & \multicolumn{2}{c}{Deit-S (ImageNet-1K)} & 22.4M & 71.3 & 83.8 & 87.7 & 80.3 & 94.8\tabularnewline

TS-CAM~\cite{DBLP:conf/iccv/GaoWPP0HZY21} &\multicolumn{2}{c}{Deit-B (ImageNet-1K)} & 87.2M & 75.8 & 84.1 & 86.6 & 86.8 & 96.7\tabularnewline

TS-CAM~\cite{DBLP:conf/iccv/GaoWPP0HZY21} & \multicolumn{2}{c}{ViT-L (LAION-2B~\cite{DBLP:journals/corr/abs-2210-08402}, ImageNet-1K, CUB(60epochs ft))}  & 304M & 63.4 & 76.0 & 80.1 & 77.3 & 93.8\tabularnewline
TS-CAM~\cite{DBLP:conf/iccv/GaoWPP0HZY21} & \multicolumn{2}{c}{ViT-H (LAION-2B~\cite{DBLP:journals/corr/abs-2210-08402}, ImageNet-1K, CUB(60epochs ft))} & 633M & 10.7 & 20.2 & 32.9 & 29.1 & 56.8\tabularnewline
\hline \rowcolor{gray0}
\rowcolor{gray0} GenPromp$\dagger$ & Stable Diffusion & EfficientNet-B7 & 1017M + 66M & \textbf{87.0} & \textbf{96.1} & \textbf{98.0} & \textbf{88.7} & \textbf{97.9}\tabularnewline
\bottomrule
\end{tabular}
    \vspace{+0.1cm}
    % \caption{Performance comparison with respect to model size and training data. With a larger backbone and pre-training dataset, the discriminatively trained method TS-CAM does not achieve higher performance. }      
    % \label{tab:param_cub}
    \end{minipage}
    \begin{minipage}{1.\linewidth}
        \centering
       \footnotesize
    \setlength{\tabcolsep}{1mm}

\begin{tabular}{ r cccccccc}
\toprule
\multirow{2}{*}{Method} & \multirow{2}{*}{Loc Back.} & \multirow{2}{*}{Cls Back.} & \multirow{2}{*}{Params.} &\multicolumn{5}{c}{ImageNet-1K} \tabularnewline
\cline{5-9}
 &   &   & &Top-1 Loc & Top-5 Loc & GT-known Loc & Top-1 Cls & Top-5 Cls\tabularnewline
\midrule

TS-CAM~\cite{DBLP:conf/iccv/GaoWPP0HZY21} & \multicolumn{2}{c}{Deit-S (ImageNet-1K)} & 22.4M & 53.4 & 64.3 & 67.6 & 74.3 & 92.1\tabularnewline
TS-CAM~\cite{DBLP:conf/iccv/GaoWPP0HZY21} & \multicolumn{2}{c}{ViT-H (LAION-2B~\cite{DBLP:journals/corr/abs-2210-08402}, ImageNet-1K(3epochs ft))} & 633M & 41.9 & 50.7 & 53.2 & 74.7 & 92.8\tabularnewline
TS-CAM~\cite{DBLP:conf/iccv/GaoWPP0HZY21} & \multicolumn{2}{c}{ViT-H (LAION-2B~\cite{DBLP:journals/corr/abs-2210-08402}, ImageNet-1K(6epochs ft))} & 633M & 42.1 & 49.9 & 52.2 & 77.4 & 93.7\tabularnewline

\hline \rowcolor{gray0}
\rowcolor{gray0} GenPromp$\dagger$ & Stable Diffusion & EfficientNet-B7 & 1017M + 66M & \textbf{65.2} & \textbf{73.4} & \textbf{75.0} & \textbf{85.1} & \textbf{97.2}\tabularnewline
\bottomrule
\end{tabular}

    \vspace{-0.1cm}
    \end{minipage} 
    \caption{Performance comparison with respect to model size and training data. With a larger backbone and pre-training dataset, the discriminatively trained method TS-CAM does not achieve higher performance.}
    \label{tab:param_sup}
\end{table*}

\begin{table*}[tbp]
    \centering
    \small
    \setlength{\tabcolsep}{1mm}

\begin{tabular}{c|cccccccc}
\toprule
&\multirow{2}{*}{Multi-resolution} & \multirow{2}{*}{Multi-timesteps} & \multirow{2}{*}{Prompt Ensemble} & \multirow{2}{*}{Prompt Embedding} & \multirow{2}{*}{Finetune} & \multicolumn{3}{c}{ImageNet-1K}\tabularnewline
\cline{7-9}
& & & & & & Top-1 Loc & Top-5 Loc & GT-known Loc\tabularnewline
\midrule
1&  & & & $f_d$ & & 58.5 & 66.0 & 67.4\tabularnewline
2&\checkmark & & & $f_d$ & & 58.6 & 66.1 & 67.5\tabularnewline
3&  & \checkmark  & & $f_d$ &  & 58.6 & 66.0 & 67.5\tabularnewline
\midrule 
4&\checkmark & \checkmark  & & $f_d$ &  & 61.2 & 69.0 & 70.4\tabularnewline
% \hline 
5&\checkmark & \checkmark  & & $f_r$ (w/o init) & & 44.6 & 50.2 & 51.3\tabularnewline

6&\checkmark & \checkmark  & &$f_r$ &  & 64.0 & 72.1 & 73.7\tabularnewline
% \hline 
7&\checkmark & \checkmark  & &$f_c$ (w/o init) &  & 56.2 & 63.2 & 64.5\tabularnewline
% \hline 
8&\checkmark & \checkmark  & & $f_c$ & & 64.5 & 72.7 & 74.2\tabularnewline
\midrule 
9&\checkmark & \checkmark  & \checkmark & $f_d$ &  & 61.5 & 69.2 & 70.7\tabularnewline

10&\checkmark & \checkmark  & \checkmark &$f_r$ & & 64.2 & 72.3 & 73.8\tabularnewline
% \hline 
11&\checkmark & \checkmark  & \checkmark & $f_c$ & & 64.6 & 72.8 & 74.3\tabularnewline
\midrule 
12&\checkmark & \checkmark  & & $f_d$ & \checkmark & 62.0 & 69.8 & 71.4\tabularnewline

13&\checkmark & \checkmark  & &$f_r$ & \checkmark & 64.9 & 73.1 & 74.6\tabularnewline
% \hline 
14&\checkmark & \checkmark  & & $f_c$ & \checkmark &65.1 & 73.3 & 74.9\tabularnewline
\midrule
15&\checkmark &  & \checkmark & $f_c$ & \checkmark & 65.0 & 73.2 & 74.8\tabularnewline
16& & \checkmark  & \checkmark & $f_c$ & \checkmark & 62.3 & 70.3 & 71.8\tabularnewline
\midrule
17&\checkmark & \checkmark  & \checkmark & $f_d$ & \checkmark & 62.2 & 70.0 & 71.5\tabularnewline

18&\checkmark & \checkmark  & \checkmark &$f_r$ & \checkmark & 64.9 & 73.1 & 74.7\tabularnewline
% \hline 
19&\checkmark & \checkmark  & \checkmark & $f_c$ & \checkmark & \textbf{65.2} & \textbf{73.4} & \textbf{75.0}\tabularnewline
\bottomrule
\end{tabular}
    \vspace{-0.1cm}
    \caption{\textbf{Ablation of main components of \Ours}. For experiments that do not have a ``$\checkmark$'' in Multi-resolution or Multi-timesteps, we use a single resolution (16$\times$ 16) or a single timestep ($t=100$) for model inference.}
       
    \label{tab:ablation}
    % \vspace{-2mm}
\end{table*}

\textbf{Additional Restults with respect to Model Size and Training Data.} In Table~\ref{tab:param_sup}, we re-implement TS-CAM with larger backbone (\eg{Deit-B, ViT-L, ViT-H}) and more training data (\eg{LAION-2B}).
As the model size getting larger, the performance of TS-CAM becomes worse on CUB-200-2011 under GT-known Loc metric, Table~\ref{tab:param_sup}(upper). As shown in Table~\ref{tab:param_sup}(lower), by finetuning ViT-H-based TS-CAM for 3 epochs on ImageNet-1K, it achieves higher classification accuracy (74.7\% $vs.$ 74.3\% on Top-1 Cls) while much lower localization accuracy (53.2\% $vs.$ 67.6\% on GT-known Loc) compared to the Deit-S-based TS-CAM.
By finetuning the model for more epochs (\eg{6 epochs}), it achievea higher classification accuracy (77.4\% $vs.$ 74.7\% on Top-1 Cls) but lower localization accuracy (52.2\% $vs.$ 53.2\% under GT-known Loc metric), demonstrating that more epochs can not improve the localization performance of TS-CAM.
We attribute this phenomenon to the inherent flaw of the discriminatively trained classification model, $i.e.$, local discriminative regions are capable of minimizing image classification loss but experience difficulty in accurate object localization. A larger backbone and more training data make this phenomenon even more serious.

\textbf{Detailed Ablation Study.}
% To better demonstrate the effectiveness of the main components in \Ours, $i.e.$, Multi-resolution, Multi-timesteps, Prompt ensemble, Prompt embedding and Finetuning, a complete table of ablation study is aggregate to Table~\ref{tab:ablation}. 
%
Table~\ref{tab:ablation} provides a detailed ablation of the performance contribution of each component and their combinations, with respect to Multi-resolution, Multi-timesteps, Prompt ensemble, Prompt embedding and Finetuning.

\section{Additional Visualization Results}

% \textbf{Comparison with Discriminatively Trained Models.}
In Fig.~\ref{fig:viz_comp_cam}, we visualize the localization results of \Ours and compared them with the discriminatively trained model (\eg{CAM}~\cite{DBLP:conf/cvpr/ZhouKLOT16}).  The object Localization maps of CAM (column b) suffer from partial object activation. Localization maps of \Ours (column d) with sole representative embeddings ($f_r$) covers more object extent but introducing background noise. Those of \Ours (column e) with combined embeddings ($f_c$) not only activate full object extent but also depress background noise for precise object localization.

We also provide additional visualization results of Fig. 2, Fig. 5 and Fig. 6 in the main document. The results are shown in Fig.~\ref{fig:viz_attn_sup}, Fig.~\ref{fig:viz_tok_sup} and Fig.~\ref{fig:viz_query_sup} respectively.

\begin{figure*}[htbp]
% 	\centering
	\includegraphics[width=0.99\linewidth]{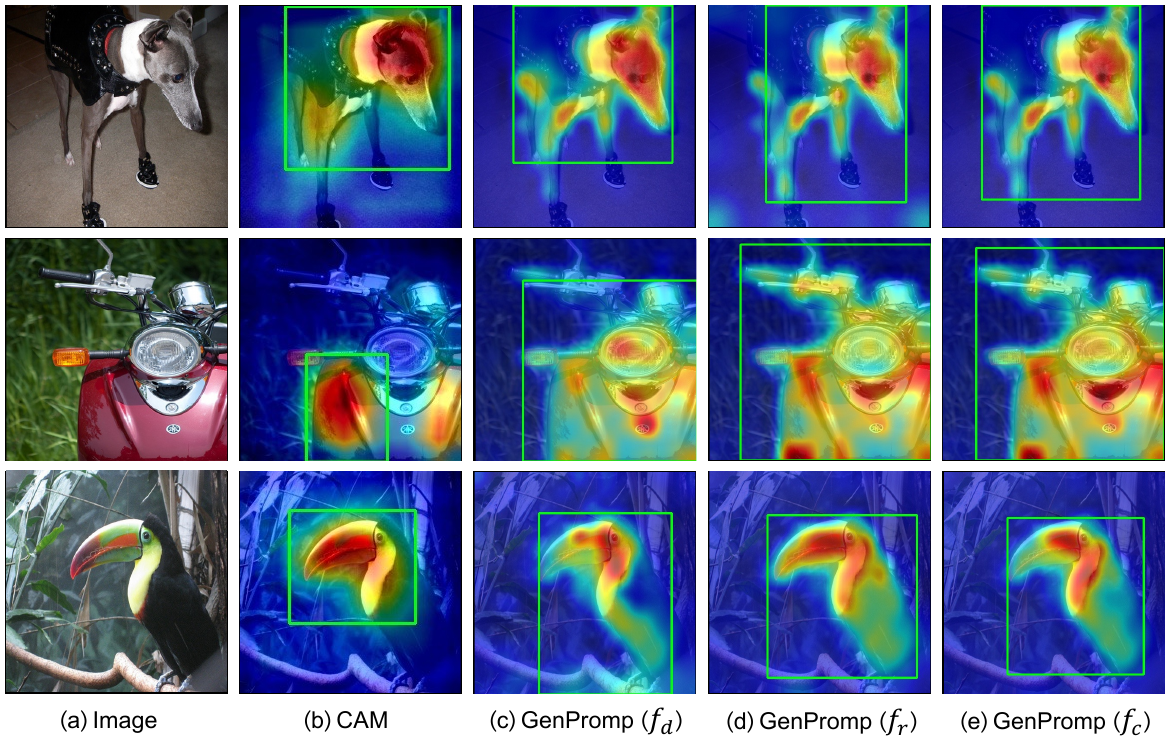}
	% \vspace{-0.2cm}
	\caption{\textbf{Comparison of activation maps between CAM and \Ours}.}
\label{fig:viz_comp_cam}
\end{figure*}

\begin{figure*}[htbp]
% 	\centering
	\includegraphics[width=0.99\linewidth]{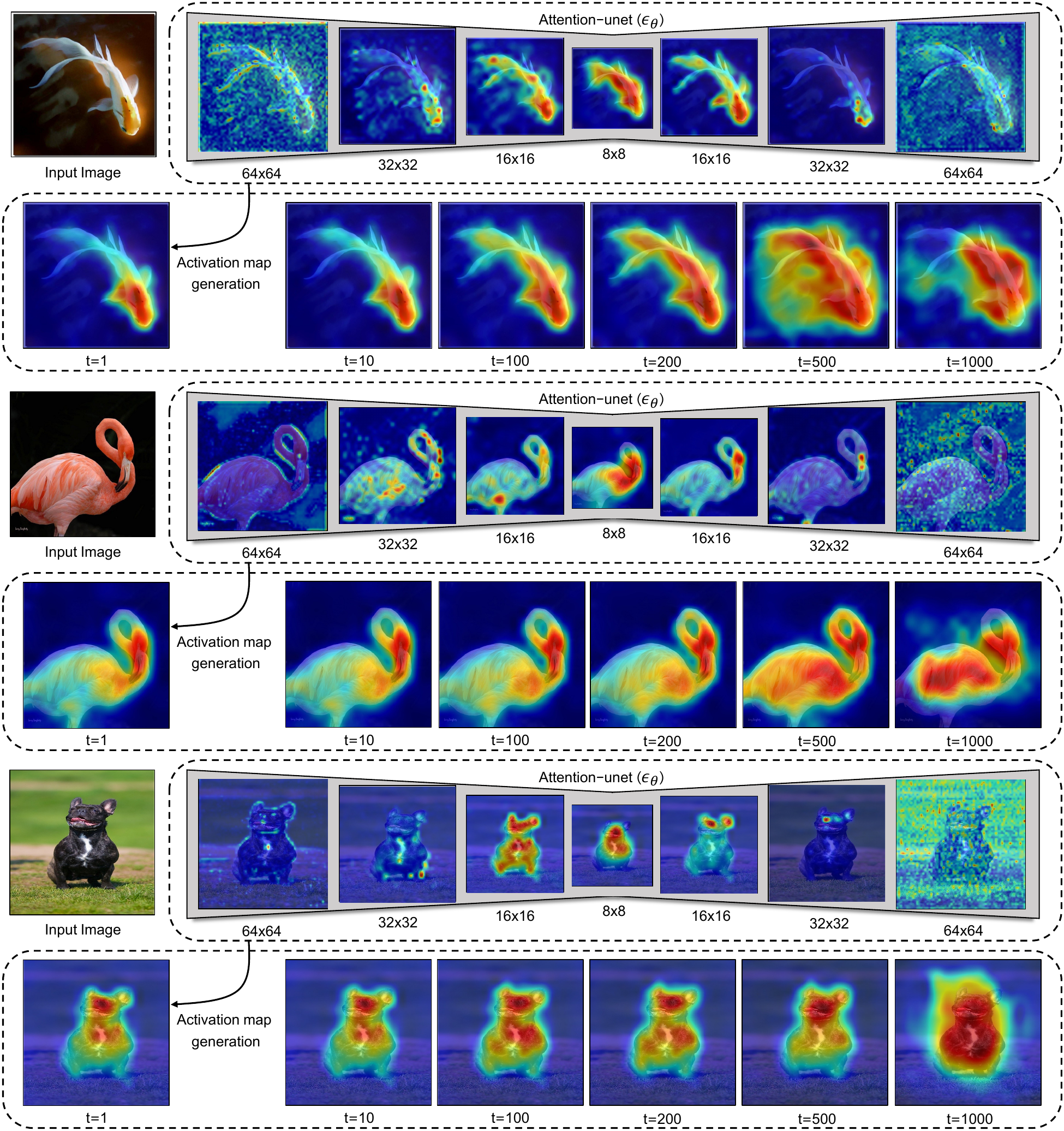}
	% \vspace{-0.2cm}
	\caption{\textbf{Visualization of cross attention maps.} Attention maps with respect to multiple resolutions and multiple noise levels (timesteps $t$) are aggregated to obtain the final localization map. The characteristics of these attention maps can be concluded as follows: (1) Attention maps with higher resolution can provide more detailed localization clues but introduce more noise. (2) Attention maps of different layers can focus on different parts of the target object. (3) Smaller $t$ provides a less noisy background but tends to partial object activation. (4) Larger $t$ activates the target object more completely but introduces more background noise.}
	\vspace{0.2cm}
\label{fig:viz_attn_sup}
\end{figure*}

\begin{figure*}[htbp]
% 	\centering
	\includegraphics[width=0.99\linewidth]{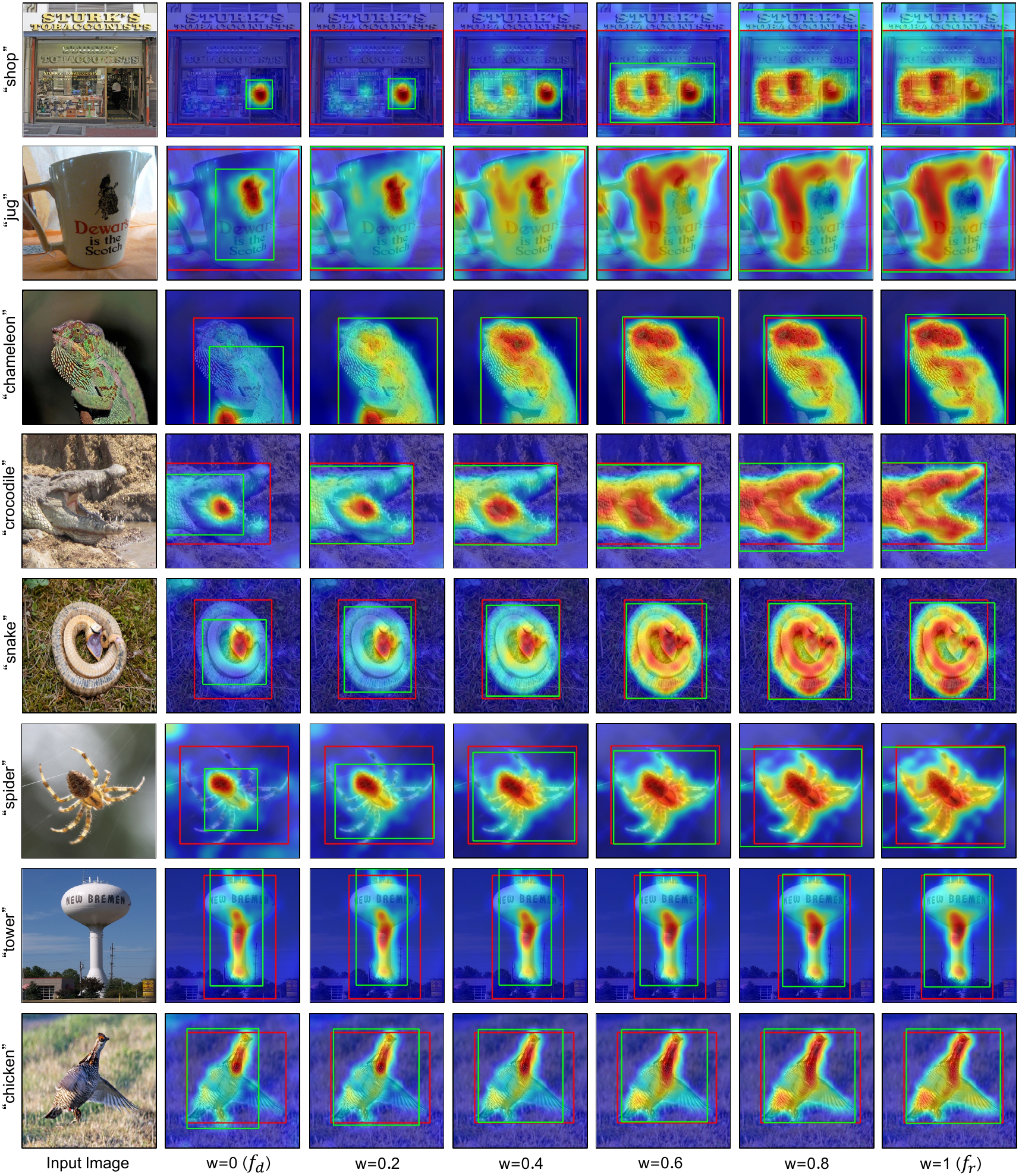}
	% \vspace{-0.2cm}
	\caption{\textbf{Activation maps and localization results using discriminative and representative embeddings.} A proper combination of discriminative embeddings $f_d$ with representative embedding $f_r$ as the prompt produces precise activation maps and good WSOL results (green boxes). }
	% \vspace{-0.2cm}
\label{fig:viz_tok_sup}
\end{figure*}

\begin{figure*}[htbp]
% 	\centering
	\includegraphics[width=0.7\linewidth]{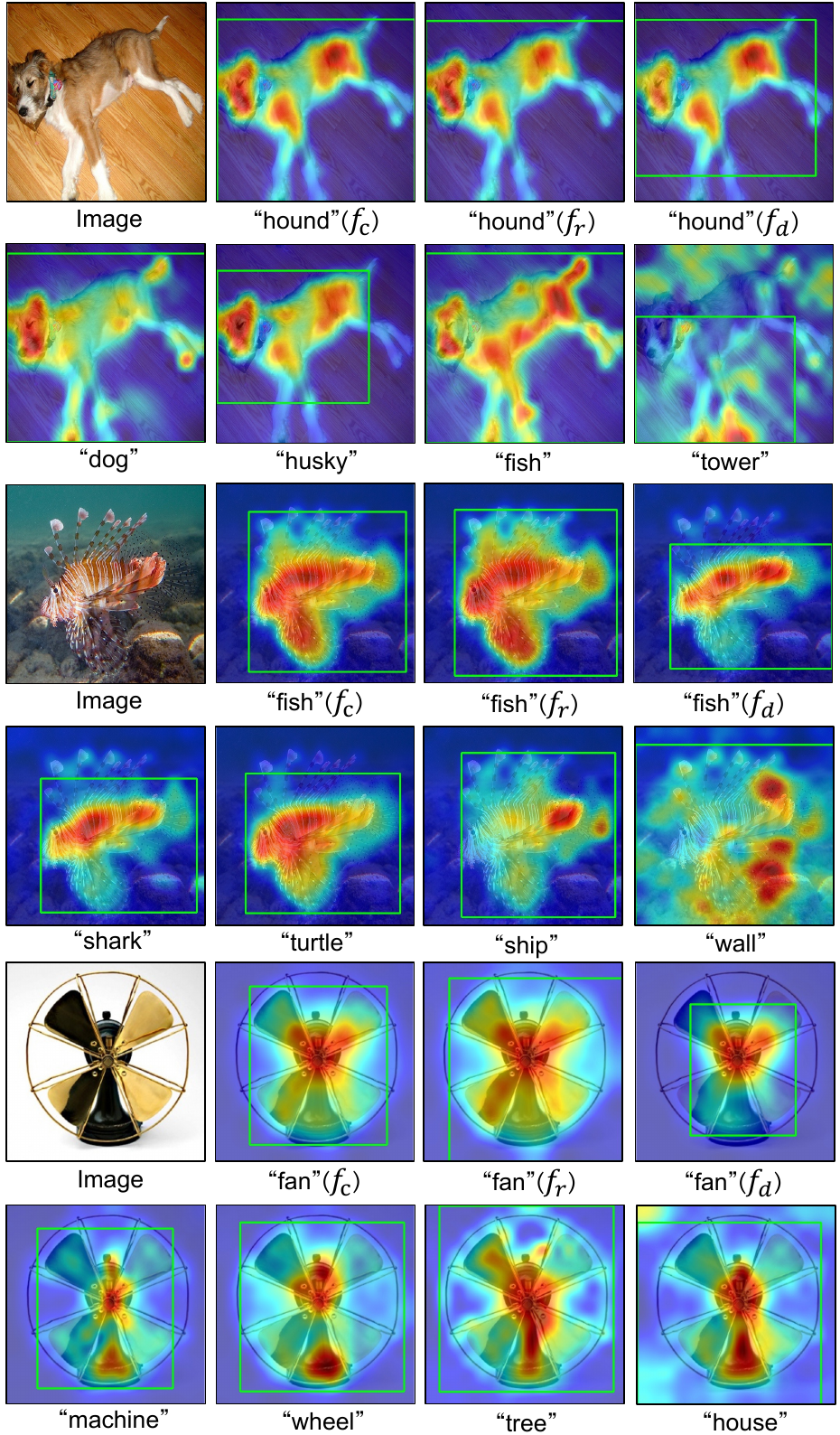}
	% \vspace{-0.2cm}
		\caption{\textbf{Object localization results of \Ours using different prompt words.} 
  % Words that similar to ``\texttt{hound}'', such as ``\texttt{dog}'',``\texttt{husky}'', can also activate the target object. Words that less related to ``\texttt{hound}'', such as ``\texttt{fish}'',``\texttt{tower}'', suffer from severe background noise and can not activate the target object.
  }
\label{fig:viz_query_sup}
\end{figure*}

\end{document}